\documentclass[a4paper,11pt]{scrartcl}
\usepackage[font=bf, large]{caption}
\usepackage[table]{xcolor}
\usepackage{amsmath}
\usepackage{amsfonts}
\usepackage{amssymb,tikz,yfonts}
\usepackage{amsbsy}
\usepackage{graphicx}
\usepackage{fancybox} 
\usepackage{fancyhdr} 
\usepackage{geometry,hhline}       
\usepackage{multicol}
\usepackage{bbm}
\usepackage{breqn}
\usepackage{multirow}
\usepackage{algorithm,algorithmic}

\usepackage[utf8x]{inputenc}
\usepackage[english]{babel}
\usepackage{float}	
\usepackage{subfig,color}
\usepackage{natbib}
\newtheorem{thm}{Theorem}[section]

\newtheorem{lemma}[thm]{Lemma}
\newtheorem{cor}[thm]{Corollary}
\newtheorem{dfn}{Definition}[section]

\newtheorem{rmk}{Remark}[section]
\allowdisplaybreaks
\newcommand{\OO}{\mathcal{O}}

\newcommand{\R}{\mathbb{R}}
\newcommand{\E}{\mathbb{E}}
\renewcommand{\P}{\mathbb{P}}

\newcommand{\eqdef}{:=}
\newcommand{\dd}{\mathrm{d}}
\newcommand{\K}{\mathcal{K}}

\newcommand{\bsig}{\boldsymbol{\sigma}}

\newcommand{\Unif}{\mathcal{U}([0,1])}

\newcommand{\bm}{\mathbf{m}}

\title{On the properties of variational approximations of
                  Gibbs posteriors}

\date{}

\author{Pierre Alquier (CREST-ENSAE), \\
 James Ridgway (CREST-ENSAE and Universit\'e Paris Dauphine)\\
and Nicolas Chopin (CREST-ENSAE and HEC Paris)}

\definecolor{Mygray}{gray}{0.85}

\begin{document}

\maketitle


\begin{abstract}
The PAC-Bayesian approach is a powerful set of techniques to derive non-asymptotic risk bounds for random estimators. The corresponding optimal
distribution of estimators, usually called the Gibbs posterior, 
is unfortunately intractable. One may sample from it using Markov chain Monte Carlo, but this is often too slow for big datasets. We consider instead variational approximations of the Gibbs posterior, which are fast to compute.  We undertake a general study of the properties of such  approximations. Our main finding is that such a variational
approximation has often the same rate of convergence as the original PAC-Bayesian procedure it approximates. 
We specialise our results to several learning tasks (classification, ranking,
matrix completion), discuss how to implement a variational approximation in
each case, and illustrate the good properties of said approximation on 
real datasets.  

\end{abstract}

\section{Introduction}
\label{sectionintroduction}

A Gibbs posterior, also known as a PAC-Bayesian or pseudo-posterior, 
is a probability distribution for random estimators of the form: 
$$ \hat{\rho}_{\lambda}({\rm d}\theta) = \frac{\exp[-\lambda r_n(\theta)]}
      {\int \exp[-\lambda r_n] {\rm d}\pi } \pi({\rm d}\theta). $$
More precise definitions will follow, but for now, $\theta$ may be 
interpreted as a parameter (in a finite or infinite-dimensional space), 
$r_n(\theta)$ as an empirical measure of risk (e.g. prediction error), 
and $\pi(\mathrm{d}\theta)$ a prior distribution. 

We will follow in this paper the PAC (Probably Approximatively Correct)-Bayesian approach, which originates from machine 
learning  \citep{Shawe-Taylor1997,McAllester1998,Catoni2004}; see~\cite{Catoni2007}
for an exhaustive study, and  \cite{Jiang2008,Yang2004, Zhang2006,Dalalyan2008} for related perspectives
(such as the aggregation of estimators in the last 3 papers). 
There, $\hat{\rho}_{\lambda}$ appears as the probability distribution that minimises the 
upper bound of an oracle inequality on the risk of \emph{random} estimators. 
The PAC-Bayesian approach offers sharp theoretical guarantees on the properties 
of such estimators, without assuming a particular model for the data generating process.

The Gibbs posterior has also appeared in other places, and under different
motivations: in Econometrics, 
as a way to avoid direct maximisation in moment estimation \citep{chernozhukov2003mcmc}; 
and in Bayesian decision theory, as as way to define a Bayesian posterior distribution
when no likelihood has been specified \citep{bissiri2013general}. 
Another well-known connection, although less directly useful 
(for Statistics), is
with thermodynamics, where $r_n$ is interpreted as an energy function, and 
$\lambda$ as the inverse of a temperature. 

Whatever the perspective, estimators derived from Gibbs posteriors usually show excellent
performance in diverse tasks, such as classification, regression, ranking, and so on, 
yet their actual implementation is still far from routine. 
The usual recommendation \citep{Dalalyan2012,Alquier2013,Guedj2013} is to sample from 
a Gibbs posterior using MCMC \citep[Markov chain Monte Carlo, see e.g.][]{GreenRobert2015}; 
but constructing an efficient MCMC sampler is often difficult, and even efficient implementations
are often too slow for practical uses when the dataset is very large. 

In this paper, we consider instead VB (Variational Bayes) approximations, which have been initially
developed to provide fast approximations of `true' posterior distributions (i.e. Bayesian posterior distributions
for a given model); see~\cite{Jordan1999,MacKay2002} and Chap. 10 in \cite{Bishop2006}.

Our main results are as follows: when PAC-Bayes bounds are available - mainly,
when a strong concentration inequality holds -  replacing the Gibbs posterior
by a variational approximation does not affect the rate of convergence to
the best possible prediction,  on the condition that the K\"ullback-Leibler
divergence between the posterior and the approximation is itself controlled
in an appropriate way.

We also provide empirical bounds, which may be computed from the data so as to ascertain
the actual performance of estimators obtained by variational approximation. 
All the results gives strong incentives, we believe, to recommend Variational Bayes
as the default approach to approximate Gibbs posteriors.

The rest of the paper is organized as follows. In Section~\ref{sectionnotation}
we introduce the notations and assumptions. In
Section~\ref{sectionVB} we introduce variational approximations and the
corresponding algorithms. The main results are provided in
general form in Section~\ref{sectionhoeffdingbernstein}:
in Subsection~\ref{sectionhoeffding}, we give results under 
the assumption that a Hoeffding type inequality holds
(slow rates) and in Subsection~\ref{sectionbernstein},
we give results under the assumption that a Bernstein type inequality
holds (fast rates). Note that for the sake of shortness, we will
refer to these settings as ``Hoeffding assumption'' and ``Bernstein
assumption'' even if this terminology is non standard.
We then apply these results in various settings: 
classification (Section~\ref{sectionclassification}), 
convex classification (Section~\ref{sec:convexclassification}), 
ranking (Section~\ref{sectionranking}),
and matrix completion (Section~\ref{sectionmatrixcompletion}). In each
case, we show how to specialise the general results of Section~\ref{sectionhoeffdingbernstein}
to the considered application, so as to obtain the properties of the VB approximation, 
and we also discuss its numerical implementation. All the proofs are collected
in the Appendix. 

\section{PAC-Bayesian framework}
\label{sectionnotation}

We observe a sample $(X_1,Y_1),\dots,(X_n,Y_n)$, taking values in $\mathcal{X}\times\mathcal{Y}$,  
where the pairs $(X_i,Y_i)$ have the same distribution $P$. We will assume explicitly that the $(X_i,Y_i)$'s are 
independent in several of our specialised results, but we do not make this assumption
at this stage, as some of our general results, and more generally the PAC-Bayesian theory,
may be extended to dependent observations; see
e.g.~\cite{AlquierLi2012}. The label set $\mathcal{Y}$ is always a subset of $\mathbb{R}$. 
A set of predictors is chosen by the statistician:
$ \{f_{\theta}: \mathcal{X}\rightarrow\mathbb{R},\,\theta\in\Theta\} .$
For example, in linear regression, we may have: $f_{\theta}(x)=\left<\theta,x\right>$,
the inner product of $\mathcal{X}=\mathbb{R}^d$, while in classification, one may have $f_{\theta}(x)=\mathbb{I}_{\left<\theta,x\right>>0}\in\{0,1\}$.

We assume we have at our disposal a risk function $R(\theta)$; typically
$R(\theta)$ is a measure of the prevision error. We set 
$\overline{R}=R(\overline{\theta})$, where $\overline{\theta}\in \arg\min_\Theta R$; i.e. $f_{\overline{\theta}}$ is an optimal predictor. 
We also assume that the risk 
function $R(\theta)$ has an empirical counterpart $r_n(\theta)$, 
and set $\overline{r}_n =r_n(\overline{\theta})$. Often, $R$ and $r_n$ 
are based on a loss function $\ell:\mathbb{R}^2\rightarrow
\mathbb{R}$; i.e. $R(\theta)=\mathbb{E}[\ell(Y,f_{\theta}(X))]$
and $\overline{r}_n(\theta)=\frac{1}{n}\sum_{i=1}^n \ell(Y_i,f_{\theta}(X_i)) $.
(In this paper, the symbol $\mathbb{E}$ will always denote the expectation with respect to the (unknown) law $P$ of the $(X_i,Y_i)$'s.)
There are situations however (e.g. ranking), where $R$ and $r_n$ have a different
form. 

We define a prior probability measure  $\pi(\cdot)$ on
the set $\Theta$ (equipped with the standard $\sigma$-algebra for the considered 
context), and we let $\mathcal{M}_+^1(\Theta)$
denote the set of all probability measures on $\Theta$.

\begin{dfn}\label{def:Gibbspost} We define, for any $\lambda>0$,
the pseudo-posterior $\hat{\rho}_{\lambda}$ by
 $$ \hat{\rho}_{\lambda}({\rm d}\theta) = \frac{\exp[-\lambda r_n(\theta)]}
      {\int \exp[-\lambda r_n] {\rm d}\pi } \pi({\rm d}\theta). $$
\end{dfn}

The pseudo-posterior $\hat{\rho}_{\lambda}$ (also known as the Gibbs
posterior, \citet{Catoni2004,Catoni2007}, or the exponentially weighted
aggregate, \citet{Dalalyan2008}) plays a central role 
in the PAC-Bayesian approach. It is obtained as the distribution that minimises the upper bound of a certain 
oracle inequality applied to \emph{random} estimators. 
Practical estimators (predictors)
may be derived from the pseudo-posterior, by e.g. taking 
the expectation, or sampling from it. 
Of course, when $\exp[-\lambda r_n(\theta)]$ 
may be interpreted as the likelihood of a certain model, 
$ \hat{\rho}_{\lambda}$ becomes a Bayesian posterior distribution, but 
we will not restrict our attention to this particular case.

The following `theoretical' counterpart of $ \hat{\rho}_{\lambda}$
will prove useful to state results.

\begin{dfn}\label{dfnpilambda} We define, for any $\lambda>0$, $\pi_{\lambda}$ as
 $$ \pi_{\lambda}({\rm d}\theta) = \frac{\exp[-\lambda R(\theta)]}
      {\int \exp[-\lambda R] {\rm d}\pi } \pi({\rm d}\theta). $$
\end{dfn}

We will derive PAC-Bayesian bounds on predictions obtained
by variational approximations of $ \hat{\rho}_{\lambda}$
under two types of assumptions:  a Hoeffding-type assumption,
from which we may deduce slow rates
of convergence (Subsection~\ref{sectionhoeffding}), and 
a Bernstein-type assumption, from which we may obtain 
fast rates of convergence 
(Subsection~\ref{sectionbernstein}).

\begin{dfn}\label{dfnhoeffding}
 We say that a Hoeffding assumption is satisfied for prior $\pi$ when there is a function
 $f$ and an interval $I\subset\R_+^*$ such that, for any $\lambda\in I$, for any $\theta\in\Theta$,
\begin{equation}
\left.
\begin{array}{l}
\pi\left(\mathbb{E} \exp\left\{ \lambda[R(\theta)-r_n(\theta)] \right\}\right) \\
\pi\left(\mathbb{E} \exp\left\{ \lambda[r_n(\theta)-R(\theta)] \right\}\right)
\end{array}
\right\}
\leq \exp\left[f(\lambda,n) \right].
\label{eq:hoeffding}
\end{equation}
\end{dfn}

Inequality \eqref{eq:hoeffding} can be interpreted as an integrated version (with respect to $\pi$) of Hoeffding's inequality, for which 
$f(\lambda,n) \asymp \lambda^2 /n$. In many cases
the loss will be bounded uniformly over $\theta$; then Hoeffding's inequality will directly imply \eqref{eq:hoeffding}. The expectation 
with respect to $\pi$ in \eqref{eq:hoeffding} allows us to treat some cases where the loss is not upper bounded by specifying a prior with sufficiently light tails.
%

\begin{dfn}\label{dfnbernstein}
 We say that a Bernstein assumption is satisfied for prior $\pi$ when there is a function
 $g$ and an interval $I\subset\mathbb{R}_+^*$ such that, for any $\lambda\in I$,
 for any $\theta\in\Theta$,
\begin{equation}
\left.
\begin{array}{l}
\pi\left(\mathbb{E} \exp\left\{ \lambda[R(\theta)-\overline{R}] - 
\lambda[r_n(\theta)-\overline{r}_n] \right\}\right) \\
\pi\left(\mathbb{E} \exp\left\{ \lambda[r_n(\theta)-\overline{r}_n] - 
\lambda[R(\theta)-\overline{R}] \right\}\right)
\end{array}
\right\}
\leq \pi\left(\exp\left[g(\lambda,n) [R(\theta)-\overline{R}] \right]\right).
\label{eq:bernstein}
\end{equation}
\end{dfn}

This assumption is
satisfied for example by sums of i.i.d. sub-exponential random variables,
see Subsection~2.4 p.~27 in~\cite{BLM2013}, when a margin assumption
on the function $R(\cdot)$ is satisfied \citep{Tsybakov2004}. This is discussed in
Section~\ref{sectionbernstein}. Again, extensions beyond the i.i.d. case
are possible, see e.g.~\cite{Wintenberger2010} for a survey and new results.
In all these examples, the important feature of the function $g$ that we will
use to derive rates of convergence is the fact that there is a constant
$c>0$ such that when $\lambda = cn$, $g(\lambda ,n)=g(cn,n) \asymp n$.

As mentioned previously, we will often consider
$r_n(\theta)=\frac{1}{n}\sum_{i=1}^n \ell(Y_i,f_{\theta}(X_i)) $,
however, the previous assumptions can also be satisfied when $r_n(\theta)$
is a U-statistic, using Hoeffding's decomposition of U-statistics
combined with the corresponding inequality for sums of independent
variables~\citep{Hoeffding1948}. This idea comes from~\cite{Clemencon2008a}
and we will use it in our ranking application.

\begin{rmk}
We could consider more generally inequalities of the form
$$
\left.
\begin{array}{l}
\pi\left(\mathbb{E} \exp\left\{ \lambda[R(\theta)-\overline{R}] - 
\lambda[r_n(\theta)-\overline{r}_n] \right\}\right) \\
\pi\left(\mathbb{E} \exp\left\{ \lambda[r_n(\theta)-\overline{r}_n] - 
\lambda[R(\theta)-\overline{R}] \right\}\right)
\end{array}
\right\}
\leq \pi\left(\exp\left[g(\lambda,n) [R(\theta)-\overline{R}]^{\kappa} \right]\right)
$$
that allow to use the more general form of the
margin assumption of \cite{Mammen1999,Tsybakov2004}.
PAC-Bayes bounds in this context are provided by~\cite{Catoni2007}. 
However, the techniques involved
would require many pages to be described so we decided to focus on 
the cases $\kappa=0$ and $\kappa=1$ to keep the exposition simple.
\end{rmk}

\section{Numerical approximations of the pseudo-posterior}
\label{sectionVB}

\subsection{Monte Carlo}
\label{sec:MC}

As already explained in the introduction, the usual approach to approximate $\hat{\rho}_{\lambda}$ is MCMC (Markov chain Monte Carlo) sampling. \cite{Ridgway2014} proposed tempering SMC (Sequential Monte Carlo, e.g. \citet{DelMoral2006}) as an alternative to MCMC to sample from Gibbs posteriors: one samples sequentially from $\hat{\rho}_{\lambda_t}$, with  $0=\lambda_0<\cdots<\lambda_T=\lambda$ where $\lambda$ 
is the desired temperature. One advantage of this approach 
is that it makes it possible to contemplate different values of $\lambda$, and choose one 
by e.g. cross-validation. Another advantage is that such an algorithm
requires little tuning; see Appendix B for more details 
on the implementation of tempering SMC. We will use tempering SMC
as our gold standard in our numerical studies. 

SMC and related Monte Carlo algorithms tend to be too slow for practical use in situations where the sample size is large, 
the dimension of $\Theta$ is large, or $f_\theta$ is expensive 
to compute. This motivates the use of fast, deterministic approximations, 
such as Variational Bayes, which we describe in the next section. 


%

\subsection{Variational Bayes}

Various versions of VB (Variational Bayes) have appeared in the literature, but the main idea is as
follows. We define a family $\mathcal{F}\subset \mathcal{M}_+^1(\Theta)$
of probability distributions that are considered as tractable. Then, we define
the VB-approximation of $\hat{\rho}_{\lambda}$: $\tilde{\rho}_{\lambda}$.

\begin{dfn}
 Let $$\tilde{\rho}_{\lambda}= \arg\min_{\rho\in\mathcal{F}} \mathcal{K}(\rho,
     \hat{\rho}_{\lambda}), $$ 
     where $\mathcal{K}(\rho,
          \hat{\rho}_{\lambda})$ denotes the KL
     (K\"ullback-Leibler) divergence of $\hat{\rho}_{\lambda}$ relative to $\rho$: $\mathcal{K}(m,\mu)
       = \int \log[\frac{{\rm d}m}{{\rm d}\mu}]{\rm d}m $ if $m\ll \mu$ (i.e. $\mu$ dominates $m$),
        $\mathcal{K}(m,\mu)=+\infty$ otherwise.
\end{dfn}

The difficulty is to find a family $\mathcal{F}$ (a) which is large enough, 
so that $\tilde{\rho}_{\lambda}$ may be close to $\hat{\rho}_{\lambda}$, 
and (b) such that computing $\tilde{\rho}_{\lambda}$ is feasible. 
We now review two types of families popular in the VB literature. 
\begin{itemize}
\item Mean field VB: for a certain decomposition 
$\Theta=\Theta_1 \times \dots \times \Theta_d$, $\mathcal{F}$
is the set of product probability measures 
\begin{equation}\label{eq:meanfield}
\mathcal{F}^{\mathrm{MF}} = \left\{\rho\in\mathcal{M}_{+}^{1}(\Theta):
  \rho({\rm d}\theta) = \prod_{i=1}^d \rho_i(\dd\theta_i),
  \forall i\in\lbrace 1,\dots,d\rbrace,\rho_i\in\mathcal{M}_+^1
  (\Theta_i)\right\}.
\end{equation}
The infimum of the KL divergence $\mathcal{K}(\rho,\hat{\rho}_\lambda)$,
relative to $\rho=\prod_i\rho_i$ satisfies 
the following fixed point condition~\citep[][Chap. 10]{Parisi1988,Bishop2006}: 
\begin{equation}\label{fixedpoint}
\forall j\in \{1,\cdots,d\}\quad\rho_j(\dd\theta_j)
\propto\exp\left(\int\left\{ -\lambda r_n(\theta)
+ \log\pi(\theta)\right\}\prod_{i\neq j}\rho_i(\dd\theta_i)\right)
\pi(\dd\theta_j). 
\end{equation}
This leads to a natural algorithm were we update successively every $\rho_j$
until stabilization.

\item Parametric family: 
$$\mathcal{F}^{\mathrm{P}} = \left\{\rho\in\mathcal{M}_{+}^{1}(\Theta):
\rho({\rm d}\theta) =  f(\theta;m)\dd\theta, m \in M  \right\}; $$
and $M$ is finite-dimensional; say $\mathcal{F}^{\mathrm{P}} $ is the family of Gaussian distributions (of dimension $d$). 
In this case, several methods may be used to compute the infimum.
As above, one may used fixed-point iteration, provided an equation 
similar to ~\eqref{fixedpoint} is available. Alternatively, one may directly
maximize $\int \log [ \exp[-\lambda r_n(\theta)]
 \frac{\dd \pi}{\dd \rho} (\theta) ]  \rho(\dd\theta) $ with respect to paramater $m$, using
numerical optimization routines. This approach was used for instance
in~\cite{Hoffman2013} with combination of some stochastic gradient descent
to perform inference on a latent Dirichlet allocation model. 
See also e.g. \cite{Khan2014,EmtiyazKhan2013} for efficient algorithms for  Gaussian variational approximation.
\end{itemize}

In what follows (Subsections~\ref{sectionhoeffding} and~\ref{sectionbernstein})
we provide tight
bounds for the prevision risk of $\tilde{\rho}_{\lambda}$. This leads to the
identification of a condition on $\mathcal{F}$ such that the risk of
$\tilde{\rho}_{\lambda}$ is not worse than the risk of $\hat{\rho}_{\lambda}$.
We will make this condition explicit in various examples, using either mean field VB
or parametric approximations.


\begin{rmk}

An useful identity, obtained by direct calculations, is: 
for any $\rho\ll \pi$,
\begin{equation}\label{equationVB}
 \log\int \exp\left[-\lambda r_n(\theta)\right] \pi(\dd\theta)
 =-\lambda \int r_n(\theta)\rho(\dd\theta)
 -\mathcal{K}(\rho,\pi)
 +\mathcal{K}(\rho,\hat{\rho}_{\lambda}).
\end{equation}
Since the left hand side does not depend on $\rho$, one sees that $\tilde{\rho}_{\lambda}$, which minimises $\mathcal{K}(\rho,\hat{\rho}_{\lambda})$
over $\mathcal{F}$, is also the minimiser of: 
 \[
	\tilde{\rho}_{\lambda} = \arg\min_{\rho\in\mathcal{F}}\left\{\int r_n(\theta)\rho(\dd\theta)+\frac1\lambda\K(\rho,\pi)\right\}
 \]
 This equation will appear frequently in the sequel in the form of an empirical upper bound.
\end{rmk}


\section{General results}
\label{sectionhoeffdingbernstein}

This section gives our general results, under either a Hoeffding Assumption (Definition \ref{dfnhoeffding}) or a Bernstein Assumption (Definition \ref{dfnbernstein}), 
on risks bounds for the variational approximation, and how it relates to risks bounds 
for Gibbs posteriors.  
These results will be specialised to several learning problems in the following sections.

\subsection{Bounds under the Hoeffding assumption}
\label{sectionhoeffding}
      
\subsubsection{Empirical bounds}


\begin{thm}
 \label{theoremhoeffding1} Under the Hoeffding assumption
 (Definition~\ref{dfnhoeffding}),
 for any $\varepsilon>0$, with probability at least $1-\varepsilon$ we have
 simultaneously for any $\rho\in\mathcal{M}_{+}^1(\Theta)$,
 $$
  \int R {\rm d}\rho
 \leq \int r_n {\rm d}\rho + \frac{ f(\lambda,n)
  + \mathcal{K}(\rho,\pi) + \log\left(\frac{1}{\varepsilon}\right)}
  {\lambda}.
 $$
\end{thm}

This result is a simple variant of a result in~\cite{Catoni2007} but
for the sake of completeness, its proof is given in
Appendix~\ref{sectionproofs}.
It gives us an upper bound on the risk of both 
the pseudo-posterior (take $\rho=\hat{\rho}_{\lambda}$) and its variational
approximation (take $\rho=\tilde{\rho}_{\lambda}$). These bounds may be be computed 
from the data, and therefore provide a simple way to evaluate the performance of 
the corresponding procedure, in the spirit of the first PAC-Bayesian  inequalities~\citep{Shawe-Taylor1997,McAllester1998,McAllester1999}.
However, this bound
do not provide the rate of convergence of these estimators. For this
reason, we also provide oracle-type inequalities.

\subsubsection{Oracle-type inequalities}

Another way to use PAC-Bayesian bounds is to
compare $\int R {\rm d}\hat{\rho}_{\lambda}$ to the best possible risk,
thus linking this approach to oracle inequalities. This is the point of
view developed in~\cite{Catoni2004,Catoni2007,Dalalyan2008}.

\begin{thm}
 \label{theoremhoeffding2} Assume that the Hoeffding assumption is satisfied (Definition~\ref{dfnhoeffding}).
  For any $\varepsilon>0$, with probability at least $1-\varepsilon$ we have
  simultaneously
 $$
  \int R {\rm d}\hat{\rho}_{\lambda}
 \leq \mathcal{B}_{\lambda}(\mathcal{M}_{+}^1(\Theta))
 := \inf_{\rho\in\mathcal{M}_{+}^1(\Theta)}
    \left\{ \int R {\rm d}\rho + 2 \frac{ f(\lambda,n)
  + \mathcal{K}(\rho,\pi) + \log\left(\frac{2}{\varepsilon}\right)}{\lambda} \right\}
 $$
 and
  $$
  \int R {\rm d}\tilde{\rho}_{\lambda}
 \leq \mathcal{B}_{\lambda}(\mathcal{F}) := \inf_{\rho\in\mathcal{F}}
    \left\{ \int R {\rm d}\rho + 2 \frac{ f(\lambda,n)
  + \mathcal{K}(\rho,\pi) + \log\left(\frac{2}{\varepsilon}\right)}{\lambda} \right\}.
 $$
 Moreover,
 $$ \mathcal{B}_{\lambda}(\mathcal{F})
 = \mathcal{B}_{\lambda}(\mathcal{M}_{+}^1(\Theta))
     + \frac{2}{\lambda} \inf_{\rho\in\mathcal{F}} \mathcal{K}(\rho,\pi_{
       \frac{\lambda}{2}}) $$
where we remind that $\pi_{\lambda}$ is defined in Definition~\ref{dfnpilambda}.
\end{thm}

In this way, we are able to compare $\int R {\rm d}\hat{\rho}_{\lambda}$ to
the best possible aggregation procedure in $\mathcal{M}_+^1(\Theta)$ and
$\int R {\rm d}\tilde{\rho}_{\lambda}$ to the best aggregation procedure in
$\mathcal{F}$. More importantly, we are able to obtain explicit 
expressions for the right-hand
side of these inequalities in various models, and thus to obtain rates of convergence.
This will be done in the remaining sections. This leads to the second
interest of this result: if there is a $\lambda=\lambda(n)$ that leads to
$\mathcal{B}_{\lambda}(\mathcal{M}_{+}^1(\Theta)) \leq \overline{R} + s_n $ with
$s_n \rightarrow
0$ for the pseudo-posterior $\hat{\rho}_{\lambda}$,
then we only have to prove that there is a $\rho\in\mathcal{F}$ such that
$ \mathcal{K}(\rho,\pi_{\lambda})/\lambda
\leq c s_n $ for some constant $c>0$ to ensure
that the VB approximation $\tilde{\rho}_{\lambda}$ also reaches the rate $s_n$.

We will see in the following sections several examples where the approximation does not deteriorate the rate of convergence. 
But first let us show the equivalent oracle inequality
under the Bernstein assumption.

\subsection{Bounds under the Bernstein assumption}
\label{sectionbernstein}

In this context the empirical bound on the risk would depend on the minimal achievable 
risk $\bar{r}_n$, and cannot be computed explicitly. We give the oracle inequality 
for both the Gibbs posterior and its VB approximation in the following theorem. 

\begin{thm}
\label{theorembernstein}
Assume that the Bernstein assumption is satisfied (Definition~\ref{dfnbernstein}).
 Assume that $\lambda>0$ satisfies $\lambda - g(\lambda,n)  > 0$. 
 Then for any $\varepsilon>0$, with probability at least $1-\varepsilon$ we have
 simultaneously:
\begin{align*}
 \int R {\rm d}\hat{\rho}_{\lambda} - \overline{R}
 & \leq \overline{\mathcal{B}}_{\lambda}\left(\mathcal{M}_{+}^1(\Theta)\right),\\
\int R {\rm d}\tilde{\rho}_{\lambda} - \overline{R} 
 & \leq \overline{\mathcal{B}}_{\lambda}(\mathcal{F}),
\end{align*}
where, for either $\mathcal{A}=\mathcal{M}_{+}^1(\Theta)$ or $\mathcal{A}=\mathcal{F}$, 
\begin{equation*}
\overline{\mathcal{B}}_{\lambda}(\mathcal{A}) = \frac{1}{\lambda - g(\lambda,n)}
 \inf_{\rho\in\mathcal{A}} \Biggl\{
 [\lambda + g(\lambda,n) ]  \int (R-\overline{R}){\rm d}\rho
  + 2 \mathcal{K}(\rho,\pi)
   + 2 \log\left(\frac{2}{\varepsilon}\right) \Biggr\}.
\end{equation*}
In addition,
$$
\overline{\mathcal{B}}_{\lambda}(\mathcal{F})
= \overline{\mathcal{B}}_{\lambda}\left(\mathcal{M}_{+}^1(\Theta)\right)
+ \frac{2}{\lambda - g(\lambda,n)} \inf_{\rho\in\mathcal{F}}
   \mathcal{K}\left(\rho,\pi_{\frac{\lambda+g(\lambda,n)}{2}}\right).
$$
\end{thm}

 The main difference with Theorem~\ref{theoremhoeffding2} is that
 the function $R(\cdot)$ is replaced by $R(\cdot)-\overline{R}$.
 This is well known way to obtain better rates of convergence.
 
\section{Application to classification}
\label{sectionclassification}

\subsection{Preliminaries}

In all this section, we assume that $\mathcal{Y}=\{0,1\}$ and we consider
linear classification: $\Theta=\mathcal{X}=\mathbb{R}^d$,
$f_{\theta}(x)=\mathbf{1}_{\left<\theta,x\right> \geq 0}$.
We put $r_{n}(\theta)=\frac{1}{n}\sum_{i=1}^{n}
   \mathbf{1}_{\{f_{\theta}(X_i)\neq Y_i\}} $, $R(\theta)
   =\mathbb{P}(Y\neq f_{\theta}(X))$
 and assume that the $[(X_i,Y_i)]_{i=1}^{n} $ are i.i.d.
In this setting, it is well-known that the Hoeffding assumption always
holds. We state as a reminder the following lemma.

\begin{lemma}
\label{lemmahoeffding1}
Hoeffding assumption~\eqref{eq:hoeffding} is satisfied with $f(\lambda,n)
 = \lambda^2 /(2n) $.
\end{lemma}
The proof is given in Appendix~\ref{sectionproofs} for the sake of completeness.

It is also possible to prove that Bernstein assumption~\eqref{eq:bernstein} holds
in the case where the so-called  margin assumption of Mammen and Tsybakov is satisfied.
This condition we use was introduced by~\cite{Tsybakov2004} in
a classification setting, based on a related definition
in~\cite{Mammen1999}.

\begin{lemma}
\label{lemmabernstein1}
 Assume that Mammen and Tsybakov's margin assumption is satisfied: i.e. there
 is a constant $C$ such that
 $$ \mathbb{E}[(\mathbf{1}_{f_{\theta}(X) \neq Y} - 
       \mathbf{1}_{f_{\overline{\theta}}(X) \neq Y})^2 ]
        \leq C[R(\theta)-\overline{R}] .$$
 Then Bernstein assumption~\eqref{eq:bernstein} is satisfied with $g(\lambda,n)
 = \frac{C\lambda^2}{2n-\lambda} $.
\end{lemma}

\begin{rmk}
 We refer the reader to~\cite{Tsybakov2004} for a proof that
 $$ \mathbb{P}(0<|\left<\overline{\theta},X\right>|\leq t)
      \leq C' t $$
 for some constant $C'>0$ implies the margin assumption. In words,
 when $X$ is not likely to be in the region $\left<\overline{\theta},X\right>
 \simeq 0$, where points are hard to classify, then the problem becomes easier and the classification rate can be improved.
\end{rmk}

We propose in this context a Gaussian prior:
$\pi = \mathcal{N}_d(0,\vartheta^2 I_d) $,
and we consider a VB approach based on Gaussian families. The corresponding  optimization problem is not convex, 
but remains feasible as we explain below.

\subsection{Three sets of Variational Gaussian approximations}

Consider the three following Gaussian families 
\begin{align*}
\mathcal{F}_1 & = \left\{\Phi_{{\bf m},\sigma^2},\,{\bf m}\in\mathbb{R}^d,
\sigma^2 \in\mathbb{R}_+^* \right\}, \\
\mathcal{F}_2 & = \left\{\Phi_{{\bf m},\bsig^2},\,{\bf m}\in\mathbb{R}^d,
\bsig^2 \in (\mathbb{R}_+^*)^2  \right\}\text{ (mean field approximation),} \\
\mathcal{F}_3 & = \left\{\Phi_{{\bf m},\Sigma},\,{\bf m}\in\mathbb{R}^d,
\Sigma \in\mathcal{S}^{d+} \right\}\text{ (full covariance approximation),}
\end{align*}
where
$\Phi_{{\bf m},\sigma^2}$ is Gaussian distribution $N_d({\bf m} ,\sigma^2 I_d)$, 
$\Phi_{{\bf m},\bsig^2}$ is $N_d(\bf m,\mathrm{diag}(\bsig^2))$,  
and 
$\Phi_{{\bf m},\Sigma}$ is $N_d(\bf m,\Sigma)$.  
Obviously, $\mathcal{F}_1\subset \mathcal{F}_2 \subset \mathcal{F}_3
 \subset \mathcal{M}_+^1 (\Theta)$, and 
\begin{equation}
\label{pleindinegalites}
\mathcal{B}_{\lambda}(\mathcal{M}_{+}^1(\Theta))
\leq \mathcal{B}_{\lambda}(\mathcal{F}_3)
\leq \mathcal{B}_{\lambda}(\mathcal{F}_2)
\leq \mathcal{B}_{\lambda}(\mathcal{F}_1).
\end{equation}

Note that, for the sake of simplicity, we will use the following
classical notations in the rest of the paper: $\varphi(\cdot)$ is
the density of $\mathcal{N}(0,1)$ w.r.t. the Lebesgue measure,
and $\Phi(\cdot)$ the corresponding c.d.f.
The rest of Section~\ref{sectionclassification} is organized as follows.
In Subsection~\ref{theory_classification}, we calculate explicitly
$\mathcal{B}_{\lambda}(\mathcal{F}_2)$
and $\mathcal{B}_{\lambda}(\mathcal{F}_1)$. Thanks to~\eqref{pleindinegalites}
this also gives an upper bound on $\mathcal{B}_{\lambda}(\mathcal{F}_3)$ and
proves the validity of the three types of Gaussian approximations. 
Then, we give details on algorithms to compute the variational approximation based on
$\mathcal{F}_2$ and $\mathcal{F}_3$, and provide a numerical illustration on real data.

\subsection{Theoretical analysis}\label{theory_classification}

We start with the empirical bound for $\mathcal{F}_2$ (and $\mathcal{F}_1$ as
a consequence),
which is a direct corollary of Theorem~\ref{theoremhoeffding1}.
\begin{cor}
For any $\varepsilon>0$, with probability at least $1-\varepsilon$ we have,
for any ${\bm}\in\mathbb{R}^d$, $\bsig^2 \in(\mathbb{R}_+)^d$,
 $$
  \int R {\rm d} \Phi_{{\bf m},\bsig^2}
 \leq \int r_n {\rm d} \Phi_{{\bf m},\bsig^2}+ \frac{ \lambda }{2n}
  + \frac{
  \sum_{i=1}^{d} \left[
  \frac{1}{2}\log\left(\frac{\vartheta^2}{\sigma_i^2}\right)
        + \frac{\sigma_i^2}{\vartheta^2}
  \right]
  + \frac{\|\bm\|^2}{\vartheta^2} - \frac{d}{2}
  +
  \log\left(\frac{1}{\varepsilon}\right)}
  {\lambda}.
 $$
\end{cor}

We now want to apply Theorem~\ref{theoremhoeffding2} in this context.
In order to do so, we introduce an additional assumption.
\begin{dfn}\label{dfnA1}
 We say that Assumption~A1 is satisfied when there is a constant $c>0$
 such that, for any $(\theta,\theta')\in\Theta^2$ with $\|\theta\|=\|\theta'\|=1$,
 $\mathbb{P}(\left<X,\theta\right>
\left<X,\theta'\right><0) \leq c \|\theta-\theta'\|$.
\end{dfn}
Note that this is not a stringent assumption. For example, it is satisfied
as soon as $X/\|X\|$ has a bounded density on the unit sphere.

\begin{cor}
\label{corclassifhoeffding}
Assume that the VB approximation is done on either $\mathcal{F}_1$,
$\mathcal{F}_2$ or $\mathcal{F}_3$.
Take $\lambda = \sqrt{nd} $ and $\vartheta = \frac1{\sqrt{d}}$.
  Under Assumption~A1, for any $\varepsilon>0$, with probability at least
  $1-\varepsilon$ we have simultaneously
 $$
 \left.
 \begin{array}{r}
  \int R {\rm d}\hat{\rho}_{\lambda} \\
  \int R {\rm d}\tilde{\rho}_{\lambda}
  \end{array}
  \right\}
 \leq
 \overline{R} + \sqrt{\frac{d}{n}} \log\left(4ne^2\right)
 + \frac{c}{\sqrt{n}}+\sqrt{\frac{d}{4n^{3}}}
 + \frac{2\log\left(\frac{2}{\varepsilon}\right)}{\sqrt{nd}}.
 $$
\end{cor}
See the appendix for a proof. Note also that the values
$\lambda=\sqrt{nd}$ and $\vartheta=\frac{1}{\sqrt{d}}$ allow to
derive this almost optimal rate of convergence, but are not
necessarily the best choices in practice.
\begin{rmk}
 Note that Assumption~A1 is not necessary to obtain oracle inequalities
 on the risk integrated under $\hat{\rho}_{\lambda}$. We refer the reader to
 Chapter 1 in~\cite{Catoni2007} for such assumption-free bounds. However, it is
 clear that without this assumption the shape of $\hat{\rho}_{\lambda}$ and
 $\tilde{\rho}_{\lambda}$ might be very different. Thus, it seems reasonable to
 require that A1 is satisfied for the approximation of $\hat{\rho}_{\lambda}$ by
 $\tilde{\rho}_{\lambda}$ to make sense.
\end{rmk}

We finally provide an application of Theorem~\ref{theorembernstein}. Under
the additional constraint that the margin assumption is satisfied, we obtain a better rate.
\begin{cor}\label{corclassifbernstein}
Assume that the VB approximation is done on either $\mathcal{F}_1$,
$\mathcal{F}_2$ or $\mathcal{F}_3$.
Under Assumption~A1 (Definition~\ref{dfnA1} page~\pageref{dfnA1}),
and under Mammen and Tsybakov margin assumption,
with $\lambda = \frac{2n}{C+2}$ and $\vartheta>0 $, for any $\varepsilon>0$,
with probability at
  least $1-\varepsilon$, 
 $$
 \left. \begin{array}{r}
  \int R {\rm d}\hat{\rho}_{\lambda} \\
  \int R {\rm d}\tilde{\rho}_{\lambda}
  \end{array} \right\}
  \leq \bar{R}+
  \frac{(C+2)(C+1)}{2}
  \left\lbrace\frac{d\log \frac n\vartheta}{n}+\frac{2d\vartheta}{n^2}+\frac2\vartheta-\frac d{\vartheta n}+\frac2{n}\log\frac{2}{\varepsilon}\right\rbrace
 +\frac{\sqrt{d} 2c(2C+1)}{n}.
 $$
\end{cor}

The prior variance optimizing the bound is $\vartheta=d/(d+2+2d/n)$,
this choice or any constant instead will lead to a rate in $d \log(n) /n$.
Note that the rate $d/n$ is minimax-optimal in this context. This is, for example,
a consequence of more general results in~\cite{Lecue} under a general form of
the the margin assumption. 
See the Appendix for a proof. 

\subsection{Implementation and numerical results}

For family $\mathcal{F}_2$ (mean field), the variational lower bound \eqref{equationVB} equals 
\begin{equation*}
  \mathcal{L}_{\lambda,\vartheta}({\bf m},\bsig)
  =-\frac\lambda{n}\sum_{i=1}^n \Phi\left(-Y_i\frac{X_i \textbf{m}}
  {\sqrt{X_{i} {\rm diag}(\bsig^2) X_{i}^t}}\right)-\frac{\bm^T\bm}{2\vartheta}+\frac12\sum_{k=1}^d\left(\log \sigma^2_k-\frac{\sigma_k^2}{\vartheta}\right),
\end{equation*}
while for family  $\mathcal{F}_3$ (full covariance), it equals 
\begin{equation*}
  \mathcal{L}_{\lambda,\vartheta}(\bm,\Sigma)=-\frac\lambda{n}\sum_{i=1}^n \Phi\left(-Y_i\frac{X_i \bm}{\sqrt{X_{i} \Sigma X_{i}^t}}\right)-\frac{\bm^T\bm}{2\vartheta}+\frac12\left(\log \vert\Sigma\vert-\frac1{\vartheta}\text{tr}\Sigma\right).
\end{equation*}

Both functions are non-convex, but the multimodality of the latter may be more severe
due to the larger dimension of  $\mathcal{F}_3$. To address this issue, we recommend to use 
the reparametrisation of \cite{Archambeau2009}, which makes the dimension of the latter optimisation problem $\OO(n)$; see \cite{Khan2014} for a related approach. In both cases, we found 
that deterministic annealing to be a good approach to optimise such non-convex functions. 
We refer to Appendix B for more details on deterministic annealing and on our particular implementation. 

%
%
%
%

We now compare the numerical performance of the mean field and full covariance VB approximations
to the Gibbs posterior (as approximated by SMC, see Section \ref{sec:MC}) for the classification
of standard datasets; see Table \ref{tab:classif}. We also include results for a kernel SVM (support vector machine); this comparison is not entirely fair, since SVM is a non-linear classifier, while all the other classifiers are linear. Still, except for the Glass dataset, the full covariance VB approximation performs as well or better than both SMC and SVM (while being much faster to compute, especially compared to SMC).

\begin{center}
\begin{table}[h]
\begin{tabular}{lccccc}
\multicolumn{1}{c}{\bf Dataset}& \multicolumn{1}{c}{\bf Covariates}&\multicolumn{1}{c}{\bf Mean Field ($\mathcal{F}_2$)} & \multicolumn{1}{c}{\bf Full cov. ($\mathcal{F}_3$)} & \multicolumn{1}{c}{\bf SMC }& \multicolumn{1}{c}{\bf SVM } 
\\ \hline \\
Pima  &7 & 31.0 & 21.3 & 22.3  &  30.4  \\
Credit & 60 & 32.0 & 33.6  & 32.0 & 32.0  \\
DNA  &180 &  23.6   &23.6 & 23.6 & 20.4  \\
SPECTF  &22 &  08.0& 06.9 & 08.5   & 10.1  \\
Glass & 10 & 34.6 & 19.6 & 23.3 &  4.7\\  
Indian & 11 & 48.0 & 25.5 & 26.2 & 26.8\\  
Breast & 10 & 35.1 & 1.1 & 1.1 & 1.7
\end{tabular}
\caption{Comparison of misclassification rates ($\%$).} 
\label{tab:classif}
\begin{minipage}{13cm}
\footnotesize{ Misclassification rates for different datasets and for the proposed approximations of the Gibbs posterior. The last column is the missclassification rate given by a kernel-SVM with radial kernel. The hyper-parameters are chosen by cross-validation.}
\vspace*{3mm}
\hrule
\end{minipage}

\end{table}
\end{center}

Interestingly, VB outperforms SMC in certain cases. This might be due to the fact that a VB
 approximation tends to be more concentrated around the mode than the Gibbs posterior it approximates.
Mean field VB does not perform so well on certain datasets (e.g. Indian). This may due either
to the approximation family being too small, or to the corresponding optmisation problem to be
strongly multi-modal.

%

\section{Application to classification under convexified loss}\label{sec:convexclassification}

Compared to the previous section, the advantage of 
convex classification is that the corresponding variational approximation will amount to minimising a convex function. This means that 
(a) the minimisation problem will be easier to deal with; 
and (b) we will be able to compute a bound for the integrated risk after a given number of steps
of the minimisation procedure. 

The setting is the same as in the previous section, except that
for convenience we now take $\mathcal{Y}=\{-1,1\}$, and the risk is based on the hinge loss,
\[
	r^H_n(\theta)=\frac1n\sum_{i=1}^n \max(0,1-Y_i<\theta,X_i>).
\]

We will write $R^H$ for the theoretical counterpart and  $\bar{R}^H$ for its minimum in $\theta$. 
We keep the superscript $H$ in order to allow comparison with the risk $R$ under the $0-1$ loss.
We assume in this section that the $X_i$ are uniformly bounded by a constant, $\vert X_i\vert<c_x$.
Note that we do not require an assumption of the form (A1) to obtain the results of this section, 
as we rely directly on the Lipschitz continuity of the hinge risk.

\subsection{Theoretical Results}
Contrarily to the previous section, the risk is not bounded in $\theta$, and 
we must specify a prior distribution for the Hoeffding assumption to hold. 

\begin{lemma}
\label{lemmahingehoeffding}
Under a independent Gaussian prior $\pi$ such that each component is $N(0,\vartheta^2)$, and 
for $\lambda<\frac{2}c\sqrt{\frac n\vartheta^2}$ and with bounded design $\vert X_{ij}\vert<c_x$,
 Hoeffding assumption~\eqref{eq:hoeffding} is satisfied with $f(\lambda,n)
= \lambda^2 /(4n)-\frac12\log\left(1-\frac{\vartheta^2\lambda^2 c_x^2}{4n}\right)$.
\end{lemma}

The main impact of such a bound is that the prior variance cannot be taken too big relative to $\lambda$. 

\begin{cor}
\label{corhinge}
Assume that the VB approximation is done on either $\mathcal{F}_1$,
$\mathcal{F}_2$ or $\mathcal{F}_3$.
Take $\lambda = \frac1{c_x}\sqrt{\frac{n}{\vartheta^2}} $ and $\vartheta = \frac1{\sqrt{d}}$.
 For any $\varepsilon>0$, with probability at least
  $1-\varepsilon$ we have simultaneously
 $$
 \left.
 \begin{array}{r}
  \int R^H {\rm d}\hat{\rho}_{\lambda} \\
  \int R^H {\rm d}\tilde{\rho}_{\lambda}
  \end{array}
  \right\}
  \leq \overline{R}^H
  +\frac{c_x}{2}\sqrt{\frac{d}n}\log\frac{n}d+2c_x\frac{d}{n}+\frac1{\sqrt{nd}}\left(\frac{c_x^2+1}{2c_x}+2c_x\log\frac2\epsilon\right)
  $$
 \end{cor}

The oracle inequality in the above corollary enjoys the same rate of convergence as the equivalent result 
in the preceding  section. In the following we link the two results.

\begin{rmk}
As stated in the beginning of the section we can use the estimator specified under the hinge loss to bound 
the excess risk of the 0-1 loss. We write $R^\star$ and $R^{H\star}$ the respective risk for their corresponding Bayes classifiers. From \cite{Zhang2004} (section 3.3) we have the following inequality, linking the excess risk under the hinge loss and the $0-1$ loss,
\[
	R(\theta)-R^\star\leq R^H(\theta)-R^{H\star}
\]
for every $\theta\in\R^p$. By integrating with respect to $\tilde{\rho}^H$ (the VB approximation on any $\mathcal{F}_1,\mathcal{F}_2,\mathcal{F}_3$ of 
the Gibbs posterior for the hinge risk) and making use of Corollary~\ref{corhinge}
we have with high probability,
\[
	\tilde{\rho}^H\left(R(\theta)\right)-R^\star\leq \inf_{\theta\in \R^p}R^H(\theta)-R^{H\star}+
	\mathcal{O}\left(\sqrt{\frac dn}\log\left(\frac nd\right)\right).
\]
\end{rmk}

\subsection{Numerical application}
We have motivated the introduction of the hinge loss as a convex upper bound. In the sequel we show 
that the resulting VB approximation also leads to a convex optimization problem. This has the advantage of opening 
a range of possible optimization algorithms \citep{Nesterov2004}. In addition we are able to bound the error of the approximated
measure after a fixed number of iterations (see Theorem~\ref{thmconvex}).

Under the model $\mathcal{F}_1$ each individual risk is given by:
\[
	\rho_{m,\sigma}(r_i(\theta))=\left(1-\Gamma_i m\right)
	\Phi\left(\frac{1-\Gamma_im}{\sigma\|\Gamma_i\|_2}\right)+
	\sigma\|\Gamma_i\|\varphi\left(\frac{1-\Gamma_im}{\sigma\|\Gamma_i\|_2}\right)
	\eqdef \Xi_i\left(\left(\!
	\begin{array}{c}
	m \\
	\sigma
	\end{array}
	\!\right)\right),
\]
writting $\Gamma_i\eqdef Y_iX_i$.

Hence the lower bound to be maximized is given by
\begin{multline*}
	\mathcal{L}(m,\sigma)=-\frac\lambda{n}\left\{\sum_{i=1}^n\left(1-\Gamma_i m\right)
	\Phi\left(\frac{1-\Gamma_im}{\sigma\|\Gamma_i\|_2}\right)+
	\sum_{i=1}^n\sigma\|\Gamma_i\|\varphi\left(\frac{1-\Gamma_im}{\sigma\|\Gamma_i\|_2}\right)\right\} \\
	-\frac{\|m\|_2^2}{2\vartheta}+\frac d2\left(\log \sigma^2 -\frac{\vartheta}{\sigma^2}\right).
\end{multline*}
It is easy to see that the function is convex in $(m,\sigma)$, first note that the map 
$$
\Psi:\left(\begin{array}{c} x \\ y\end{array}\right)
\mapsto x\Phi\left(\frac xy\right)+y\varphi\left(\frac xy\right),
$$
is convex and note that we can write $\Xi_i\left(\left(\!
	\begin{array}{c}
	m \\
	\sigma
	\end{array}
	\!\right)\right)=\Psi\left(A\left(\begin{array}{c} x \\ y\end{array}\right)+b\right)$ hence by 
	composition of convex function with linear mappings we have the result. 
	Similar reasoning could be held for the case $\mathcal{F}_2$ and $\mathcal{F}_3$, where in
later the parametrization should be done in C such that $\Sigma=CC^t$. The bound is however not universally Lipschitz in 
$\sigma$, this impacts the optimization algorithms.

On the class of function $\mathcal{F}_0=\left\{\Phi_{m,\frac1n},m\in\mathbb{R}^d\right\}$, for which our 
Oracle inequalities still hold we could get faster numerical algorithms. The objective function has Lipschitz 
continuous derivatives and we would get a rate of $\frac{L}{(1+k)^2}$.

Other convex loss could be considered which could lead to convex optimization problems. For instance one could
consider the exponential loss.

\begin{table}[h]
\begin{center}
\begin{tabular}{lccc}
	\multicolumn{1}{c}{\bf Dataset}& \multicolumn{1}{c}{\bf Covariates}&\multicolumn{1}{c}{\bf Hinge loss} &\multicolumn{1}{c}{\bf SMC} 
\\ \hline \\
Pima  &7 & 21.8 & 22.3\\
Credit & 60 & 27.2 & 32.0\\
DNA  &180 &  4.2 &  23.6 \\
SPECTF  &22 &  19.2 & 08.5 \\
Glass & 10 & 26.12 & 23.3\\  
Indian & 11 & 26.2 &  25.5\\  
Breast & 10 & 0.5 & 1.1
\end{tabular}
\end{center}
\caption{Comparison of misclassification rates ($\%$).} 
\label{sample-table}
\begin{minipage}{13cm}
\footnotesize{ Misclassification rates for different datasets and 
for the proposed approximations of the Gibbs posterior. The hyperparameters are chosen by cross-validation. This is to be compared to Table \ref{tab:classif}.}
\vspace*{3mm}
\hrule
\end{minipage}

\end{table}

\begin{thm}
\label{thmconvex}
Assume that the VB approximation is done on $\mathcal{F}_1,\mathcal{F}_2 \text{ or } \mathcal{F}_3$. Denote by $\tilde{\rho}_k(\dd\theta)$ 
the VB approximated measure after the $k$th iteration of an optimal convex solver using the hinge loss. 
Take $\lambda=\sqrt{nd}$ and $\vartheta=\frac1{\sqrt{d}}$ then under the hypothesis of Corollary \ref{corhinge}
with probability $1-\epsilon$
\[
	 \int R^H \dd \tilde{\rho}_k\leq \overline{R}^H+\frac{LM}{\sqrt{1+k}}
	    + 
  +\frac{c_x}{2}\sqrt{\frac{d}n}\log\frac{n}d+2c_x\frac{d}{n}+\frac1{\sqrt{nd}}\left(\frac{c_x^2+1}{2c_x}+2c_x\log\frac2\epsilon\right)
  \]
where $L$ is the Lipschitz coefficient on a ball of radius $M$ of the
objective function maximized in VB.
\end{thm}

From Theorem~\ref{thmconvex} we can compute the number of iterations to get a given 
level of error at a given probability.

We find that on average the misclassification error (Table \ref{sample-table}) is lower than for the 0-1 loss
where we have no guaranties that the maximum is attained.

\section{Application to ranking}
\label{sectionranking}

\subsection{Preliminaries}

In this section we take $\mathcal{Y}=\{0,1\}$ and  consider again 
linear classifiers: $\Theta=\mathcal{X}=\mathbb{R}^d$,
$f_{\theta}(x)=\mathbf{1}_{\left<\theta,x\right> \geq 0}$. We
consider however a different criterion: in ranking, not only
we want to classify well an object $x$, but we want to make sure that given
two different objects, the one that is more likely to correspond to a label $1$
will be assigned a larger score through the function $f_{\theta}$. A usual way
to measure this is to introduce the risk function
$$
R(\theta)
   =\mathbb{P}[(Y_1-Y_2)(f_{\theta}(X_1)-f_{\theta}(X_2))<0]
$$
and the empirical risk
$$
r_{n}(\theta)=\frac{1}{n(n-1)}\sum_{1\leq i \neq j \leq n}
   \mathbf{1}_{\{(Y_i-Y_j)(f_{\theta}(X_i)-f_{\theta}(X_j))<0\}}.
$$

Then, again, we recall classical results.
\begin{lemma}
\label{lemmahoeffding2}
  The Hoeffding-type assumption is satisfied with $f(\lambda,n)
 = \frac{\lambda^2}{n-1} $.
\end{lemma}

The variant of the margin assumption adapted to ranking was established by~\cite{Robbiano2013} and \cite{Ridgway2014}.
\begin{lemma}
\label{lemmabernstein2}
Assume the following
 margin assumption:
  $$ \mathbb{E}[(\mathbf{1}_{[f_{\theta}(X_1)-f_{\theta}(X_2)][Y_1-Y_2]<0}
  - \mathbf{1}_{[f_{\overline{\theta}}(X_1)-f_{\overline{\theta}}(X_2)][Y_1-Y_2]<0}
    )^2]\leq C[R(\theta)-\overline{R}] .$$
 Then Bernstein assumption~\eqref{eq:bernstein} is satisfied with $g(\lambda,n)
 = \frac{C\lambda^2}{n-1-4\lambda} $.
\end{lemma}

We still consider a Gaussian prior
$$\pi({\rm d}\theta) =
\prod_{i=1}^d \varphi(\theta_i;0,\vartheta^2) {\rm d}\theta_i $$
and the approximation families will be the same as in
Section~\ref{sectionclassification}:
$
\mathcal{F}_1  = \{\Phi_{{\bf m},\sigma^2},{\bf m}\in\mathbb{R}^d,
\sigma^2 \in\mathbb{R}_+^* \}$,
$
\mathcal{F}_2  = \{\Phi_{{\bf m},\bsig^2},{\bf m}\in\mathbb{R}^d,
\bsig^2 \in (\mathbb{R}_+^*)^2  \}$ and
$
\mathcal{F}_3  = \{\Phi_{{\bf m},\Sigma},{\bf m}\in\mathbb{R}^d,
\Sigma \in\mathcal{S}^{d+} \}$.

\subsection{Theoretical study}

Here again, we start with the empirical bound.
\begin{cor}
	\label{corempirical}
For any $\varepsilon>0$, with probability at least $1-\varepsilon$ we have,
for any $\bm\in\mathbb{R}^d$, $\sigma^2 \in(\mathbb{R}_+)^d$,
 $$
  \int R {\rm d} \Phi_{{\bf m},\bsig^2}
 \leq \int r_n {\rm d} \Phi_{{\bf m},\bsig^2} + \frac{ \lambda }{n-1}
  + \frac{
  \sum_{j=1}^{d} \left[
  \frac{1}{2}\log\left(\frac{\vartheta^2}{\sigma_i^2}\right)
        + \frac{\sigma_i^2}{\vartheta^2}
  \right]
  + \frac{\|\bm\|^2}{\vartheta^2} - \frac{d}{2}
  +
  \log\left(\frac{1}{\varepsilon}\right)}
  {\lambda}.
 $$
\end{cor}

In order to derive a theoretical bound, we introduce the following variant
of Assumption~A1.
\begin{dfn}
 We say that Assumption~A2 is satisfied when there is a constant $c>0$
 such that, for any $(\theta,\theta')\in\Theta^2$ with $\|\theta\|=\|\theta'\|=1$,
$\mathbb{P}(\left<X_1-X_2,\theta\right>
\left<X_1-X_2,\theta'\right><0) \leq c \|\theta-\theta'\|$.
\end{dfn}
Assumption~A2 is satisfied as soon as $(X_1-X_2)/\|X_1-X_2\|$ has a bounded
density on the unit sphere.

\begin{cor}\label{corrankinghoeffding}
Use either $\mathcal{F}_1$, $\mathcal{F}_2$ or $\mathcal{F}_3$.
 Take $\lambda = \sqrt{\frac{d(n-1)}{2}} $ and $\vartheta = 1$.
  Under (A2), for any $\varepsilon>0$, with probability at least $1-\varepsilon$, 
 $$
\left.
 \begin{array}{r}
  \int R {\rm d}\hat{\rho}_{\lambda} \\
  \int R {\rm d}\tilde{\rho}_{\lambda}
  \end{array}
  \right\}
 \leq
 \overline{R} + \sqrt{\frac{2d}{n-1}} \left(1
  + \frac{1}{2}\log\left(2d(n-1)\right)\right) + \frac{c\sqrt{2}}{\sqrt{n-1}}
 + \frac{2\sqrt{2}\log\left(\frac{2{\rm e}}{\varepsilon}\right)}{\sqrt{(n-1)d}}.
 $$
\end{cor}

Finally, under an additional margin assumption, we have:
\begin{cor}
  Under Assumption~A2 and the margin assumption of Lemma~\eqref{lemmabernstein2},
  for
  $\lambda = \frac{n-1}{C+5} $ and $\vartheta >0 $,
  for any $\varepsilon>0$, with probability at least $1-\varepsilon$, 
\begin{multline*}
\left. \begin{array}{r}
  \int R {\rm d}\hat{\rho}_{\lambda} \\
  \int R {\rm d}\tilde{\rho}_{\lambda}
  \end{array} \right\}
  \leq \bar{R}+
  \frac{(C+5)(C+1)}{2}\left\lbrace\frac{d\log \frac n\vartheta}{n-1}
  +\frac{2d\vartheta}{n(n-1)}+\frac2\vartheta-\frac d{\vartheta n-1}
  +\frac2{n-1}\log\frac{2}{\varepsilon}\right\rbrace
  \\
  +\frac{\sqrt{d} 4c(C+1)}n.
\end{multline*}
\end{cor}

The prior variance optimizing the bound is $\vartheta=d/(d+2+2d/n)$.
The proof is similar to the ones of
Corollaries~\ref{corclassifhoeffding},~\ref{corclassifbernstein}
and~\ref{corrankinghoeffding}.

As in the case of classification, ranking under an AUC loss can be done 
by replacing the indicator function by the corresponding upper bound given by 
an hinge loss. In this case we can derive similar results as for the convexified classification 
in particular we can get a convex minimization problem and obtain result without requiring assumption
(A2).

\subsection{Algorithms and numerical results}

As an illustration we focus here on family $\mathcal{F}_2$ (mean field).
In this case the VB objective to maximize is given by:
\begin{equation}
	\mathcal{L}(\bm,\sigma^2)=-\frac\lambda{n_+ n_-}\sum_{i:y_i=1,j:y_j=0} \Phi\left(-\frac{\Gamma_{ij} m}{\sqrt{\sum_{k=1}^d(\gamma^k_{ij})^2\sigma_k^2}}\right)
	-\frac{\|\bm\|^2_2}{2\vartheta}+\frac12\sum_{k=1}^d \left[\log\sigma^2_k-\frac{\sigma^2_k}{\vartheta}\right],
\end{equation}
where $\Gamma_{ij}=X_i-X_j$, and  where $(\gamma^k_{ij})_k$ are the elements of $\Gamma$.

This function is expensive to compute, as it involves $n_{+}n_{-}$ terms, the computation of which is $\mathcal{O}(p)$.

We propose to use a stochastic gradient descent in the spirit of \cite{Hoffman2013}. The model we consider is not in an exponential family, meaning we cannot use the trick developed by these authors. We propose instead to use a standard descent. 

The idea is to replace the gradient by a unbiased version based on a batch of size $B$ as described in Algorithm \ref{algo-SGD}
in the Appendix.  
\cite{Robbins1951} show that for a step-size $(\lambda_t)_t$ such that $\sum_t\lambda_t^2<\infty$ and $\sum_t\lambda_t=\infty$ the algorithm converges to a local optimum. 

In our case we propose to sample pairs of data with replacement and use the unbiased version of the derivative of the risk component.
We use a simple gradient descent without any curvature information. One could also use recent research on stochastic 
quasi Newton-Raphson \citep{Byrd2014}.

For illustration, we consider a small dataset (Pima), and a larger one (Adult). The latter is already 
quite challenging with $n_+n_-=193,829,520$ pairs to compare. In both cases with different size of batches 
convergence is obtained with a few iterations only and leads to acceptable bounds. 

\begin{figure}[t]
\begin{center}
\begin{tabular}{lll}
\subfloat[Pima]{\includegraphics[scale=0.4]{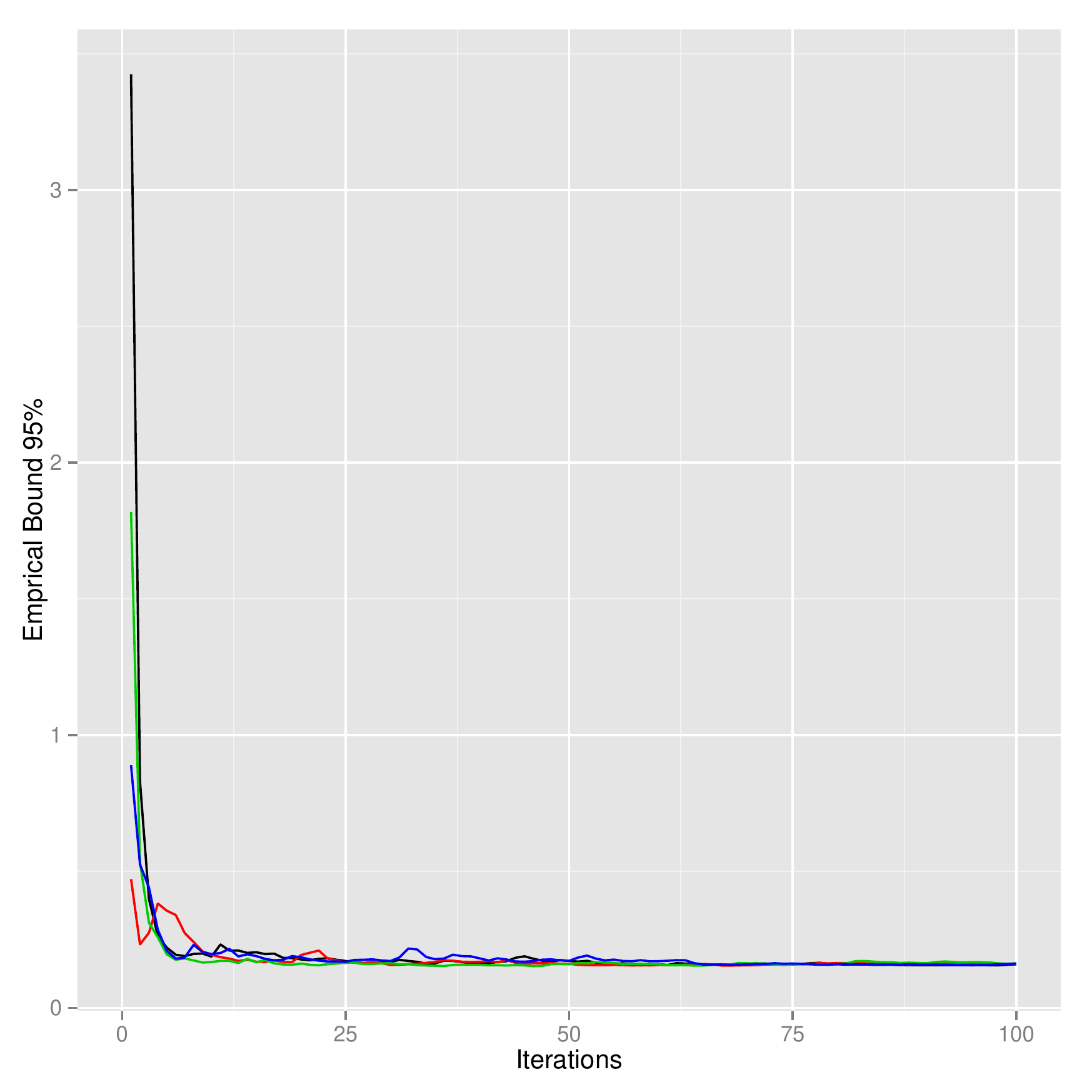}} &
\subfloat[adult]{\includegraphics[scale=0.4]{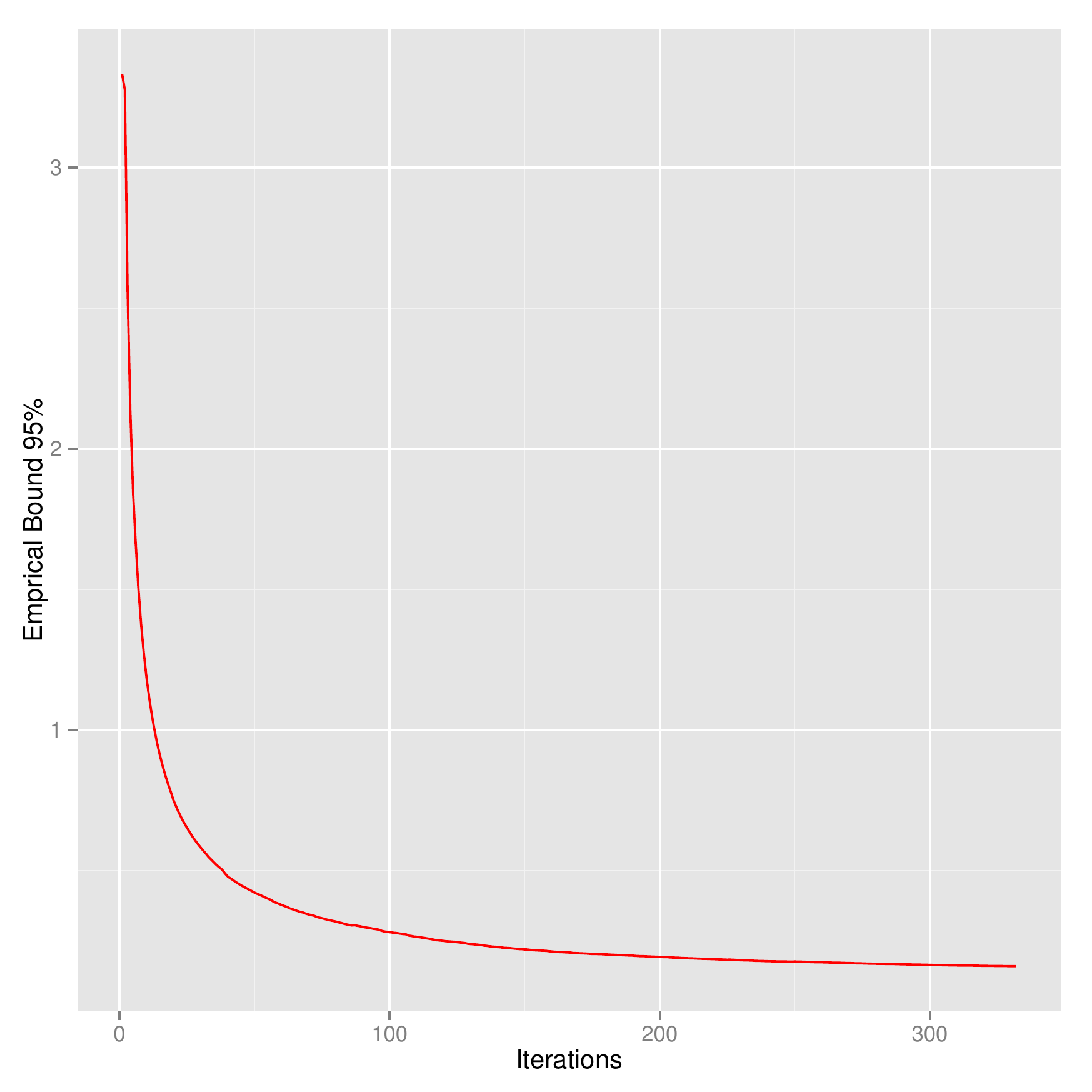} } &
\end{tabular}
\begin{minipage}{15cm}
\caption{Error bound at each iteration, stochastic descent, Pima and Adult datasets.}
\label{fig:SVB}
\footnotesize{Stochastic VB with fixed temperature $\lambda=100$ for Pima and $\lambda=1000$ for adult. The left panel shows several curves
that correspond to different batch sizes; these curves are hard to distinguish. 
The right panel is for a batch size of 50. 
The adult dataset has $n=32556$ observation and $n_+n_-=193829520$ possible pairs. The convergence is obtained in order of seconds. 
The bounds are the empirical bounds obtained in Corollary ~\ref{corempirical} for a probability of 95\%.
}
\vspace*{3mm}
\hrule
\end{minipage}
\end{center}
\end{figure}

In Figure \ref{fig:SVB} we show the empirical bound on the AUC risk as a function of the iteration of 
the algorithm, for several batch sizes. The bound is taken for $95\%$ probability, the batch sizes are taken to be 
$B=1,10,20,50$ for the Pima dataset, and 50 for the Adult dataset.  The figure shows an additional feature
of VB approximation in the context of Gibbs posterior: namely the possibility of computing the empirical upper 
bound given by Corollary~\ref{corempirical}. That is we can check the quality of the bound at each iteration of the algorithm,
or for different values of the hyperparameters.

\section{Application to matrix completion}
\label{sectionmatrixcompletion}
The matrix completion problem has received increasing attention recently, partly
due to spectacular theoretical results \citep{candes2010power},
and to challenging applications like the Netflix challenge \citep{Bennett2007}.
In the perspective of this paper,
the specific interest of this application is twofold.
First, this is a case where the
family of approximations is not parametric, but rather of the form \eqref{eq:meanfield}, 
i.e. the family of products of independent
components. 
Then, there is no known theoretical result for the Gibbs estimator
in the considered model,
yet we can still directly bound the 
loss induced by the variational approximation. 

We observe
i.i.d. pairs $((X_i,Y_i))_{i=1}^n$ where $X_i \in\{1,\dots,m_1\}\times
\{1,\dots,m_2\}$, and we assume that there is a
$m_1\times m_2$-matrix $M$ such that $Y_i=M_{X_i}+\varepsilon_i$ and
the $\varepsilon_i$ are centred. Assuming that $X_i$ is uniform
on $\{1,\dots,m_1\}\times\{1,\dots,m_2\}$, that $f_{\theta}(X_i) = \theta_{X_i} $, and taking the quadratic risk,
$R(\theta)=\mathbb{E}\left[(Y_i-\theta_{X_i})^2\right]$,
we have that 
$$ R(\theta)-\overline{R} = \frac{1}{m_1 m_2}\|\theta-M\|_F^2 $$
where $\|\cdot\|_F$ stands for the Frobenius norm.

A common way to parametrise the problem is 
$$ \Theta = \{\theta= UV^T, U\in \mathbb{R}^{m_1 \times K},
   V\in\mathbb{R}^{m_2 \times K}\} $$
where $K$ is large; e.g. $K=\min(m_1,m_2)$. 
Following \cite{salakhutdinov2008}, we define the following prior 
distribution:
$ U_{\cdot,j} \sim \mathcal{N}(0,\gamma_j I)  $,
$ V_{\cdot,j} \sim \mathcal{N}(0,\gamma_j I)  $
where the $\gamma_j$'s are i.i.d. from an inverse gamma
distribution, $\gamma_j \sim \mathcal{I}\Gamma(a,b)$.

Note that VB algorithms were used in this context by \cite{Lim2007} (with a slightly simpler prior however: the $\gamma_j$'s
are fixed rather than random). Since then, this prior and
variants were used in several papers \citep[e.g.][]{lawrence2009,zhou2010}. Until now,
no theoretical results were proved up to our knowledge.
Two papers prove minimax-optimal rates for slightly modified estimators (by truncation),
for which
efficient algorithms are unknown
\citep{Mai2014,Suzuki2014}. However, using Theorems~\ref{theoremhoeffding2}
and~\ref{theorembernstein} we are able to prove the following: \emph{if}
there is a PAC-Bayesian bound leading to a rate for $\hat{\rho}_{\lambda}$ in
this context, then the same rate holds for $\tilde{\rho}_{\lambda}$. In other
words: if someone proves the conjecture that the Gibbs estimator is minimax-optimal
(up to $\log$ terms) in this context, then the VB approximation will enjoy automatically
the same property.

We propose the following approximation:
$$
\mathcal{F} = \left\{
\rho({\rm d}(U,V)) =
       \prod_{i=1}^{m_1} u_{i}({\rm d} U_{i,\cdot}) 
       \prod_{j=1}^{m_2} v_{j}({\rm d} V_{j,\cdot})
\right\}.
$$

\begin{thm}
 Assume that $M=UV^T$ with $|U_{i,k}|,|V_{j,k}|\leq C$. Assume that
 ${\rm rank}(M) = r $ so that we can assume that $U_{\cdot,r+1}
  = \dots = U_{\cdot,K} = V_{\cdot,r+1} = \dots = V_{\cdot,K} = 0$
 (note that the prior $\pi$ does not depend on the knowledge of $r$ though).
 Choose the prior distribution on the hyper-parameters $\gamma_j$ as
 inverse gamma $\mathrm{Inv-}\Gamma (a,b)$ with 
 $b\leq 1/[2\beta (m_1\vee m_2)
 \log(2 K (m_1\vee m_2))]$.
 Then there is a constant $\mathcal{C}(a,C)$ such that,
 for any $\beta>0$,
 $$ \inf_{\rho\in\mathcal{F}} \mathcal{K}(\rho,\pi_{\beta})
   \leq
\mathcal{C}(a,C) \left\{ r(m_1 + m_2 ) \log\left[\beta b (m_1+m_2)K\right]
 +\frac{1}{\beta} \right\}.
$$
\end{thm}

See the Appendix for a proof. 

For instance, in Theorem~\ref{theorembernstein}, in classification 
and ranking we had $\lambda$, $\lambda-g(\lambda,n)$ and
$\lambda+g(\lambda,n)$ of order $\OO(n)$. In this case we would
have:
$$
\frac{2}{\lambda - g(\lambda,n)} \inf_{\rho\in\mathcal{F}}
   \mathcal{K}\left(\rho,\pi_{\frac{\lambda+g(\lambda,n)}{2}}\right)
   = \mathcal{O} \left( 
 \frac{\mathcal{C}(a,C) r(m_1 + m_2 ) \log\left[n b (m_1+m_2)K\right] }{n}
   \right),
$$
and note that in this context it is know that the minimax rate is
at least
$r(m_1+m_2)/n$ \citep{koltchinskii2011}.

\subsection{Algorithm}

As already mentioned, the approximation family is not parametric in this case, but 
rather of type mean field. The corresponding VB algorithm amounts to iterating equation~\eqref{fixedpoint}, 
which takes the following form in this particular case: 
\begin{align*}
	u_j(\dd U_{j,.})&\propto \exp\left\{-\frac\lambda n 
		\sum_{i}\E_{V,U_{-j}}\left[(Y_{X_i}-(UV^T)_{X_i})^2\right]-\sum_{k=1}^K\E_{\gamma_j}\left[\frac1{2\gamma_k}\right]U^2_{jk}\right\}\\
	v_j(\dd V_{j,.})&\propto \exp\left\{-\frac\lambda n 
		\sum_{i}\E_{V_{-j},U}\left[(Y_{X_i}-(UV^T)_{X_i})^2\right]-\sum_{k=1}^K\E_{\gamma_j}\left[\frac1{2\gamma_k}\right]V^2_{jk}\right\}\\
	p(\gamma_k)&\propto \exp\left\{-\frac1{2\gamma_k}
	\left(\sum_j \E_U U^2_{kj}+\sum_i \E_V V^2_{ik}\right)+(\alpha+1)\log \frac1{\gamma_k}-\frac\beta{\gamma_k} \right\}
\end{align*}
where the expectations are taken with respect to the thus defined variational approximations. One recognises Gaussian distributions for the first two, and an
inverse Gamma distribution for the third. We refer to \cite{Lim2007}
for more details on this algorithm and for a numerical illustration. 

\section{Discussion}

We showed in several important scenarios that approximating a Gibbs posterior through VB (Variational Bayes) techniques does not deteriorate the rate of convergence of the corresponding procedure. We also described practical algorithms for fast computation 
of these VB approximations, and provided empirical bounds that may be computed from 
the data to evaluate the performance of the so-obtained VB-approximated procedure. 
We believe these results provide a strong incentive to recommend VB as the default
approach to approximate Gibbs posteriors, in lieu of Monte Carlo methods. 

We hope to extend our results to other applications beyond those discussed in this paper,
such as regression. One technical difficulty with regression is that the risk
function is not bounded, which makes our approach a bit less direct to apply. 
In many papers on PAC-Bayesian bounds for regression, the noise can be unbounded
(usually, it is assumed to be sub-exponential),
but one assumes that the predictors are bounded, see e.g. \cite{Alquier2013}.
However, using the robust loss function of Audibert and Catoni,
it is possible to relax this assumption
\citep{Audibert2011,Catoni2012}. This requires a more technical
analysis, which we leave for further work. 
%
%

\bibliography{biball}

\appendix

\section{Proofs}
\label{sectionproofs}

\subsection{Preliminary remarks}

We start by a general remark.
Let $h$ be a function $\Theta \rightarrow \mathbb{R}_+$
with $\int \exp[-h(\theta)] \pi(\dd \theta)< \infty$.
Let us put
$$\pi[h](\dd \theta) = \frac{\exp[-h(\theta)]}
{\int \exp[-h(\theta')] \pi(\dd \theta')} \pi(\dd \theta). $$
Direct calculation yields, for any $\rho \ll \pi$
with $\int h {\rm d}\rho < \infty$,
\begin{equation*}
\mathcal{K}(\rho,\pi[h])
 = \lambda \int h {\rm d}\rho
  + \mathcal{K}(\rho,\pi) + \log \int \exp(-h){\rm d}\pi.
\end{equation*}
Two well known consequences are
\begin{align*}
\pi[h] & = \arg\min_{\rho\in\mathcal{M}_{+}^1(\Theta)}
 \left\{  \int h {\rm d}\rho
  + \mathcal{K}(\rho,\pi) \right\},   \\
- \log \int \exp(-h){\rm d}\pi 
& = \min_{\rho\in\mathcal{M}_{+}^1(\Theta)}
 \left\{   \int h {\rm d}\rho
  + \mathcal{K}(\rho,\pi) \right\}.
\end{align*}
We will use these inequalities many times in the followings.
The most frequent application will be with $h(\theta)=\lambda
r_n (\theta)$ (in this case $\pi[\lambda r_n] = \hat{\rho}_\lambda$)
or $h(\theta)=\pm \lambda
[r_n (\theta) - R(\theta)]$, the first case leads to
\begin{align}
\label{dv}
\mathcal{K}(\rho,\hat{\rho}_{\lambda})
 & = \lambda \int r_n{\rm d}\rho
  + \mathcal{K}(\rho,\pi) + \log \int \exp(-\lambda r_n){\rm d}\pi,
  \\
\hat{\rho}_{\lambda} & = \arg\min_{\rho\in\mathcal{M}_{+}^1(\Theta)}
 \left\{ \lambda  \int r_n {\rm d}\rho
  + \mathcal{K}(\rho,\pi) \right\},  \label{optim} \\
- \log \int \exp(-\lambda r_n){\rm d}\pi 
& = \min_{\rho\in\mathcal{M}_{+}^1(\Theta)}
 \left\{ \lambda  \int r_n {\rm d}\rho
  + \mathcal{K}(\rho,\pi) \right\}. \label{optim2}
\end{align}
We will use~\eqref{dv},~\eqref{optim} and~\eqref{optim2} several
times in this appendix.

\subsection{Proof of the theorems in Subsection~\ref{sectionhoeffding}}

\noindent \textit{Proof of Theorem~\ref{theoremhoeffding1}.}
This proof follows the standard PAC-Bayesian approach (see~\cite{Catoni2007}).
Apply Fubini's theorem to the first inequality of~\eqref{eq:hoeffding}:
 $$
\mathbb{E} \int \exp\left\{ \lambda[R(\theta)-r_n(\theta)] -f(\lambda,n)\right\}
\pi({\rm d}\theta)
\leq 1
$$
then apply the preliminary remark with $h(\theta)=
\lambda [r_n (\theta)-R(\theta)]$: 
 $$
\mathbb{E}  \exp\left\{ \sup_{\rho\in\mathcal{M}_{+}^1(\Theta) } \int
\lambda[R(\theta)-r_n(\theta)] \rho({\rm d}\theta) - \mathcal{K}(\rho,\pi)
 - f(\lambda,n) \right\}
\leq 1.
$$
Multiply both sides by $\varepsilon$ and use $\mathbb{E}[\exp (U)]
\geq \mathbb{P}(U>0)$ for any $U$ to obtain:
$$
\mathbb{P}\left[
\sup_{\rho\in\mathcal{M}_{+}^1(\Theta) } \int
\lambda[R(\theta)-r_n(\theta)] \rho({\rm d}\theta) - \mathcal{K}(\rho,\pi)
 - f(\lambda,n) + \log(\varepsilon)>0
\right] \leq \varepsilon.
$$
Then consider the complementary event:
$$
\mathbb{P}\left[ \forall \rho\in\mathcal{M}_{+}^1(\Theta),
\quad
 \lambda \int R {\rm d}\rho 
 \leq \lambda  \int r_n {\rm d}\rho + f(\lambda,n)
  + \mathcal{K}(\rho,\pi) + \log\left(\frac{1}{\varepsilon}\right)
  \right] \geq 1-\varepsilon.
$$
$\square$

\noindent \textit{Proof of Theorem~\ref{theoremhoeffding2}.}
Using the same calculations as above, 
we have, with probability at least
$1-\varepsilon$, simultaneously for all $\rho\in\mathcal{M}_{+}^{1}(\Theta)$,
\begin{align}
\label{pacb1}
 \lambda \int R {\rm d}\rho 
& \leq \lambda  \int r_n {\rm d}\rho + f(\lambda,n)
  + \mathcal{K}(\rho,\pi) + \log\left(\frac{2}{\varepsilon}\right)
  \\
\label{pacb2} 
 \lambda \int r_n  {\rm d}\rho 
 & \leq \lambda  \int R {\rm d}\rho + f(\lambda,n)
  + \mathcal{K}(\rho,\pi) + \log\left(\frac{2}{\varepsilon}\right).
\end{align}
We use~\eqref{pacb1} with $\rho=\hat{\rho}_{\lambda}$ and~\eqref{optim} to get
$$
 \lambda \int R {\rm d}\hat{\rho}_{\lambda}
 \leq \inf_{\rho\in\mathcal{M}_{+}^1 (\Theta)}
  \left\{\lambda  \int r_n {\rm d}\rho + f(\lambda,n)
  + \mathcal{K}(\rho,\pi) + \log\left(\frac{2}{\varepsilon}\right)\right\}
$$
and plugging~\eqref{pacb2} into the right-hand side, we obtain
$$
 \lambda \int R {\rm d}\hat{\rho}_{\lambda}
 \leq \inf_{\rho\in\mathcal{M}_{+}^1 (\Theta)}
  \left\{\lambda  \int R {\rm d}\rho + 2 f(\lambda,n)
  + 2 \mathcal{K}(\rho,\pi) + 2 \log\left(\frac{2}{\varepsilon}\right)\right\}.
$$
Now, we work with $\tilde{\rho}_{\lambda}=\arg\min_{\rho\in\mathcal{F}}
\mathcal{K}(\rho,\hat{\rho}_{\lambda})$. Plugging~\eqref{dv}
into~\eqref{pacb1} we get, for any $\rho$,
$$
 \lambda \int R {\rm d}\rho
 \leq  f(\lambda,n)
   + \mathcal{K}(\rho,\hat{\rho}_{\lambda}) 
  -\log \int \exp(-\lambda r_n){\rm d}\pi
  +\log\left(\frac{2}{\varepsilon}\right).
$$
By definition of $\tilde{\rho}_{\lambda}$, we have:
$$
 \lambda \int R {\rm d}\tilde{\rho}_{\lambda}
 \leq \inf_{\rho\in\mathcal{F}} \left\{
   f(\lambda,n) + \mathcal{K}(\rho,\hat{\rho}_{\lambda}) 
  -\log \int \exp(-\lambda r_n){\rm d}\pi
  +\log\left(\frac{2}{\varepsilon}\right)\right\}
$$
and, using~\eqref{dv} again, we obtain:
$$
 \lambda \int R {\rm d}\tilde{\rho}_{\lambda}
 \leq \inf_{\rho\in\mathcal{F}} \left\{
\lambda  \int r_n {\rm d}\rho + f(\lambda,n)
  +  \mathcal{K}(\rho,\pi)
 + \log\left(\frac{2}{\varepsilon}\right)\right\}.
$$
We plug~\eqref{pacb2} into the right-hand side to obtain:
$$
 \lambda \int R {\rm d}\tilde{\rho}_{\lambda}
 \leq \inf_{\rho\in\mathcal{F}} \left\{
\lambda  \int R {\rm d}\rho + 2 f(\lambda,n)
  +  2 \mathcal{K}(\rho,\pi)
 + 2 \log\left(\frac{2}{\varepsilon}\right)\right\}.
$$
This proves the second inequality of the theorem. In order to prove
the claim
$$ \mathcal{B}_{\lambda}(\mathcal{F}) =
   \mathcal{B}_{\lambda}(\mathcal{M}_{+}^1 (\Theta)) + \frac{2}{\lambda}
    \inf_{\rho\in\mathcal{F}}
     \mathcal{K}(\rho,\pi_{\frac{\lambda}{2}}), $$
note that
\begin{align*}
\mathcal{B}_{\lambda}(\mathcal{F})
& = \inf_{\rho\in\mathcal{F}} \left\{
  \int R {\rm d}\rho + \frac{2 f(\lambda,n)}{\lambda}
  +  \frac{2 \mathcal{K}(\rho,\pi)}{\lambda}
 + \frac{2 \log\left(\frac{2}{\varepsilon}\right)}{\lambda}\right\}
\\
& =  \inf_{\rho\in\mathcal{F}} \left\{
 -\frac{2}{\lambda} \log \int \exp\left(-\frac{\lambda}{2}R \right){\rm d}\pi
  +  \frac{2 f(\lambda,n)}{\lambda}
  +  \frac{2 \mathcal{K}(\rho,\pi_{\frac{\lambda}{2}})}{\lambda}
 + \frac{2 \log\left(\frac{2}{\varepsilon}\right)}{\lambda}\right\}
\\
& = 
 -\frac{2}{\lambda} \log \int \exp\left(-\frac{\lambda}{2}R \right){\rm d}\pi
  +  \frac{2 f(\lambda,n)}{\lambda}
 + \frac{2 \log\left(\frac{2}{\varepsilon}\right)}{\lambda}
 + \frac{2}{\lambda}\inf_{\rho\in\mathcal{F}}
  \mathcal{K}(\rho,\pi_{\frac{\lambda}{2}})
 \\
 & =  \mathcal{B}_{\lambda}(\mathcal{M}_{+}^1 (\Theta)) + \frac{2}{\lambda}
    \inf_{\rho\in\mathcal{F}}
     \mathcal{K}(\rho,\pi_{\frac{\lambda}{2}}).
\end{align*}
This ends the proof.
$\square$

\subsection{Proof of Theorem~\ref{theorembernstein} (Subsection~\ref{sectionbernstein})}

\noindent \textit{Proof of Theorem~\ref{theorembernstein}.}
As in the proof of Theorem~\ref{theoremhoeffding1}, we apply Fubini, then~\eqref{optim2} to the first inequality of~\eqref{eq:bernstein} to obtain
\begin{equation*}
\mathbb{E}  \exp\left\{ \sup_{\rho} \int \left[ \lambda[R(\theta)-\overline{R}] - 
\lambda[r_n(\theta)-\overline{r}_n] - g(\lambda,n) [R(\theta)-\overline{R}]
 \right] \rho({\rm d}\theta) - \mathcal{K}(\rho,\pi)
 \right\} 
 \leq 1
\end{equation*}
and we multiply both sides by $\varepsilon/2$ to get
\begin{equation} \label{step1thmb}
\mathbb{P}\Biggl\{
\sup_{\rho} \Biggl[
[\lambda-g(\lambda,n)] \left[ \int R {\rm d}\rho - \overline{R} \right]
 \geq \lambda \left[ \int r_n {\rm d}\rho - \overline{r}_n \right]
  + \mathcal{K}(\rho,\pi) + \log\left(\frac{2}{\varepsilon}\right)
\Biggr]
\Biggr\} 
\leq \frac{\varepsilon}{2}.
\end{equation}
We now consider the second inequality in~\eqref{eq:bernstein}:
$$
\mathbb{E} \exp\left\{ \lambda[r_n(\theta)-\overline{r}_n]
 -\lambda[R(\theta)-\overline{R}] 
- g(\lambda,n) [R(\theta)-\overline{R}] \right\} \leq 1.
$$
The same derivation leads to
\begin{equation} \label{step2thmb}
\mathbb{P}\Biggl\{
\sup_{\rho} \Biggl[
[\lambda-g(\lambda,n)] \left[ \int r_n {\rm d}\rho - \overline{r}_n \right]
 \geq
 \lambda  \left[ \int R {\rm d}\rho - \overline{R} \right]
  + \mathcal{K}(\rho,\pi) + \log\left(\frac{2}{\varepsilon}\right)
\Biggr]
\Biggr\} 
\leq \frac{\varepsilon}{2}.
\end{equation}
We combine~\eqref{step1thmb} and~\eqref{step2thmb} by a union bound argument,
and we consider the complementary event: with probability at least
$1-\varepsilon$, simultaneously for all $\rho\in\mathcal{M}_{+}^{1}(\Theta)$,
\begin{equation}\label{step3thmb}
[\lambda-g(\lambda,n)] \left[ \int R {\rm d}\rho - \overline{R} \right]
 \leq \lambda \left[ \int r_n {\rm d}\rho - \overline{r}_n \right]
  + \mathcal{K}(\rho,\pi) + \log\left(\frac{2}{\varepsilon}\right),
\end{equation}
\begin{equation}\label{step4thmb}
\lambda \left[ \int r_n {\rm d}\rho - \overline{r}_n \right]
 \leq [\lambda+g(\lambda,n)]\left[ \int R {\rm d}\rho - \overline{R} \right]
  + \mathcal{K}(\rho,\pi) + \log\left(\frac{2}{\varepsilon}\right).
\end{equation}
We now derive consequences of these two inequalities (in other words, we
focus on the event where these two inequalities are satisfied).
Using~\eqref{optim} in~\eqref{step3thmb} yields
\begin{equation*}
[\lambda-g(\lambda,n)]
\left[ \int R {\rm d}\hat{\rho}_{\lambda} - \overline{R} \right]
 \leq
 \inf_{\rho\in\mathcal{M}_{+}^1(\Theta)} \left\{
 \lambda \left[ \int r_n {\rm d}\rho - \overline{r}_n \right]
  + \mathcal{K}(\rho,\pi)
  + \log\left(\frac{2}{\varepsilon}\right) \right\}.
\end{equation*}
We plug~\eqref{step4thmb} into the right-hand side to obtain:
\begin{multline*}
[\lambda-g(\lambda,n)]
\left[ \int R {\rm d}\hat{\rho}_{\lambda}-\overline{R} \right]
 \\ \leq
 \inf_{\rho\in\mathcal{M}_{+}^1(\Theta)} \Biggl\{
 [\lambda+g(\lambda,n)]\left[ \int R {\rm d}\rho - \overline{R} \right]
  + 2 \mathcal{K}(\rho,\pi)
  + 2 \log\left(\frac{2}{\varepsilon}\right) \Biggr\}.
\end{multline*}
Now, we work with $\tilde{\rho}_{\lambda}$.
Plugging~\eqref{dv} into~\eqref{step1thmb} we get
\begin{equation*}
[\lambda-g(\lambda,n)] \left[ \int R {\rm d}\rho - \overline{R} \right]
 \leq \mathcal{K}(\rho,\hat{\rho}_{\lambda})
 - \log \int \exp[-\lambda (r_n-\overline{r}_n)]{\rm d}\pi
 + \log\left(\frac{2}{\varepsilon}\right).
\end{equation*}
By definition of $\tilde{\rho}_{\lambda}$, we have:
\begin{multline*}
[\lambda-g(\lambda,n)] \left[ \int R {\rm d}\tilde{\rho}_{\lambda}
- \overline{R} \right]
 \\
\leq \inf_{\rho\in\mathcal{F}} \left\{ \mathcal{K}(\rho,\hat{\rho}_{\lambda})
 - \log \int \exp[-\lambda (r_n-\overline{r}_n)]{\rm d}\pi
 + \log\left(\frac{2}{\varepsilon}\right) \right\}.
\end{multline*}
Then, apply~\eqref{dv} again to get:
$$
[\lambda-g(\lambda,n)] \left[ \int R {\rm d}\tilde{\rho}_{\lambda}
- \overline{R} \right]
 \leq \inf_{\rho\in\mathcal{F}} \left\{
 \lambda \int (r_n-\overline{r}_n){\rm d}\rho
    + \mathcal{K}(\rho,\pi)
 + \log\left(\frac{2}{\varepsilon}\right) \right\}.
$$
Plug~\eqref{step4thmb} into the right-hand side to get
\begin{multline*}
[\lambda-g(\lambda,n)]
\left[ \int R {\rm d}\tilde{\rho}_{\lambda} - \overline{R} \right]
 \\
 \leq \inf_{\rho\in\mathcal{F}} \left\{
 [\lambda+g(\lambda,n)]\int (R-\overline{R}){\rm d}\rho
    + 2 \mathcal{K}(\rho,\pi)
 + 2 \log\left(\frac{2}{\varepsilon}\right) \right\}.
\end{multline*}
$\square$

\subsection{Proofs of Section~\ref{sectionclassification}}

\noindent \textit{Proof of Lemma~\ref{lemmahoeffding1}.}
Combine
Theorem 2.1 p. 25 and Lemma 2.2 p. 27 in~\cite{BLM2013}.
$\square$

\noindent \textit{Proof of Lemma~\ref{lemmabernstein1}.}
Apply Theorem 2.10 in~\cite{BLM2013}, and plug the margin assumption.
$\square$

\noindent \textit{Proof of Corollary~\ref{corclassifhoeffding}.}
We remind that thanks to~\eqref{pleindinegalites} it is enough to prove
the claim for $\mathcal{F}_1$.
We apply Theorem~\ref{theoremhoeffding2} to get:
\begin{align*}
 \mathcal{B}_{\lambda}(\mathcal{F}_1) & = \inf_{({\bf m},\sigma^2)}
    \left\{ \int R {\rm d}\Phi_{{\bf m},\sigma^2} + \frac{\lambda}{n} + 2 \frac{
  \mathcal{K}(\Phi_{{\bf m},\sigma^2},\pi) + \log\left(\frac{2}{\varepsilon}\right)}{\lambda} \right\}
  \\
  & = \inf_{(m,\sigma^2)}
    \left\{ \int R {\rm d}\Phi_{{\bf m},\sigma^2} + \frac{\lambda}{n} + 2 \frac{
d \left[
  \frac{1}{2}\log\left(\frac{\vartheta^2}{\sigma^2}\right)
        + \frac{\sigma^2}{\vartheta^2}
  \right]
  + \frac{\|\bm\|^2}{\vartheta^2} - \frac{d}{2}
 + \log\left(\frac{2}{\varepsilon}\right)}{\lambda} \right\}.
\end{align*}
Note that the minimizer of $R$, $\overline{\theta}$, is not
unique (because $f_{\theta}(x)$ does not depend on $\|\theta\|$) and we
can chose it in such a way that $\|\overline{\theta}\|=1$. Then
\begin{align*}
R(\theta) - \overline{R}
 & = \mathbb{E} \left[ \mathbf{1}_{\left<\theta,X\right>Y<0}
  -\mathbf{1}_{\left<\overline{\theta},X\right>Y<0} \right]
 \leq \mathbb{E} \left[ \mathbf{1}_{\left<\theta,X\right>
   \left<\overline{\theta},X\right> <0} \right]
   \\
 & = \mathbb{P}\left(\left<\theta,X\right>
   \left<\overline{\theta},X\right> <0\right)
 \leq c \left\|\frac{\theta}{\|\theta\|}-\overline{\theta}\right\|
\leq 2 c \|\theta-\overline{\theta}\|.
\end{align*}
So:
\begin{multline*}
\mathcal{B}_{\lambda}(\mathcal{F}_1)
    \leq \overline{R}
    + \inf_{({\bf m},\sigma^2)}
    \biggl\{  2c \int \|\theta-\overline{\theta}\|\Phi_{{\bf m},\sigma^2}({\rm d}\theta)
    \\
    + \frac{\lambda}{n} + 2 \frac{
d \left[
  \frac{1}{2}\log\left(\frac{\vartheta^2}{\sigma^2}\right)
        + \frac{\sigma^2}{\vartheta^2}
  \right]
  + \frac{\|\bm\|^2}{\vartheta^2} - \frac{d}{2}
 + \log\left(\frac{2}{\varepsilon}\right)}{\lambda} \biggr\}.
\end{multline*}
We now restrict the infimum to distributions $\nu$ such that $\bm=\overline{\theta}$:
$$
\mathcal{B}(\mathcal{F}_1)
    \leq \overline{R}
    + \inf_{\sigma^2}
    \left\{  2c \sqrt{d} \sigma
    + \frac{\lambda}{n} + \frac{
  d\log\left(\frac{\vartheta^2}{\sigma^2}\right)
        + \frac{2d \sigma^2}{\vartheta^2}
  + \frac{2}{\vartheta^2} - d
 + 2\log\left(\frac{2}{\varepsilon}\right)}{\lambda} \right\}.
$$
We put $\sigma=\frac{1}{2\lambda}$ and substitute $\frac1{\sqrt{d}}$ for $\vartheta$ to get
$$
\mathcal{B}(\mathcal{F}_1)
    \leq \overline{R}
    + \frac{\lambda}{n} + \frac{ c\sqrt{d} 
    + d\log(4\frac{\lambda^2}{d})+\frac{d^{2}}{2\lambda^2}+d
 + 2\log\left(\frac{2}{\varepsilon}\right)}{\lambda}.
$$
Substitute $\sqrt{nd}$ for $\lambda$
to get the desired result.
$\square$

\noindent \textit{Proof of Corollary~\ref{corclassifbernstein}.}
We apply Theorem~\ref{theorembernstein}:
\begin{multline*}
	\int (R-\overline{R}) {\rm d}\tilde{\rho}_\lambda
\\
\leq\inf_{{\bf m},\sigma^2}\left\lbrace\frac{\lambda+g(\lambda,n)}
{\lambda-g(\lambda,n)}\int (R-\bar{R})\dd\Phi_{{\bf m},\sigma^2}
+\frac1{\lambda-g(\lambda,n)} \left(2\mathcal{K}(\Phi_{{\bf m},\sigma^2},\pi)
+2\log\frac2\epsilon\right)\right\rbrace
\end{multline*}
where $\lambda<\frac{2n}{C+1}$.  Computations similar to those in the
the proof of Corollary~\ref{corclassifhoeffding} lead to
\begin{multline*}
\int R d\tilde{\rho}_\lambda
    \leq \overline{R}
 + \inf_{{\bf m},\sigma^2}
    \Biggl\{  2c\frac{\lambda+g(\lambda,n)}{\lambda-g(\lambda,n)}
    \int \|\theta-\overline{\theta}\|\Phi_{{\bf m},\sigma^2}({\rm d}\theta)
    \\
    + 2 \frac{
\sum_{j=1}^{d} \left[
  \frac{1}{2}\log\left(\frac{\vartheta^2}{\sigma^2}\right)
        + \frac{\sigma^2}{\vartheta^2}
  \right]
  + \frac{\|\bm\|^2}{\vartheta^2} - \frac{d}{2}
 + \log\left(\frac{2}{\varepsilon}\right)}{\lambda-g(\lambda,n)} \Biggr\}.
\end{multline*}
taking $\bm=\bar{\theta}$ and $\lambda=\frac{2n}{C+2}$, we get the result.
$\square$

\subsection{Proofs of Section~\ref{sec:convexclassification}}

\noindent \textit{Proof of Lemma~\ref{lemmahingehoeffding}.}
For fixed $\theta$ we can upper bound the individual risk such that:
\[
	0\leq \max(0,1-<\theta,X_i>Y_i)\leq 1+\vert<\theta,X_i>\vert
\]
such that we can apply Hoeffding's inequality conditionally on $X_i$ and fixed $\theta$.

We get,
\begin{align*}
	\E\left[\exp\left(\lambda(R^H-r^H_n)\right)\vert X_1,\cdots,X_n\right]&\leq \exp\left\{\frac{\lambda^2}{8n^2}\sum_{i=1}^n (1+\vert<\theta,X_i>\vert)^2\right\}\\
	&\leq \exp\left\{\frac{\lambda^2}{4 n}+\frac{\lambda^2 c_x^2}{4n}\|\theta\|^2\right\}
\end{align*}
where the last inequality stems from the fact that $\left(a+b\right)^2\leq2\left(a^2+b^2\right)$ and the fact that we have supposed the $X_i$ to be bounded. We can take the expectation of this term with respect to the
$X_i$'s and with respect to our Gaussian prior.
\begin{align*}
	\pi\left\{\E\left[\exp\left(\lambda(R^H-r^H_n)\right)\right]\right\}&\leq \frac{\exp\left(\frac{\lambda^2}{4n}\right)}{(2\pi)^{\frac d2}\sqrt{\vartheta^2}}\int \exp\left(\frac{\lambda^2c_x^2}{4 n}\|\theta\|^2-\frac1{2\vartheta^2}\|\theta\|^2\right)\dd\theta\\
	& \leq  \frac{\exp\left(\frac{\lambda^2}{4n}\right)}{(2\pi)^{\frac d2}\sqrt{\vartheta^2}}\int \exp\left(-\frac12\left[\frac1{\vartheta^2}-\frac{\lambda^2c_x^2}{4 n}\right]\|\theta\|^2\right)\dd\theta
\end{align*}
The integral is a properly defined Gaussian integral under the hypothesis that $\frac1{\vartheta^2}-\frac{\lambda^2c_x^2}{4 n}>0$ hence $\lambda<\frac{2}{c_x}\sqrt{\frac n\vartheta^2}$. The integral is proportional to a Gaussian and we can directly write:
\[
	\pi\left\{\E\left[\exp\left(\lambda(R^H-r^H_n)\right)\right]\right\}\leq\frac{\exp\left(\frac{\lambda^2}{4n}\right)}{\sqrt{1-\frac{\vartheta^2\lambda^2 c_x^2}{4 n}}}
\]
writing everything in the exponential gives the desired result.
$\square$

\noindent \textit{Proof of Corollary~\ref{corhinge}.}
We apply Theorem~\ref{theoremhoeffding2} to get:
\begin{dmath*}
 \mathcal{B}_{\lambda}(\mathcal{F}_1)  = \inf_{({\bf m},\sigma^2)}
 \left\{ \int R^H {\rm d}\Phi_{{\bf m},\sigma^2} + \frac{\lambda}{2n}-\frac1{\lambda}\log\left(1-\frac{\vartheta^2\lambda^2 c_x^2}{4n}\right) + 2 \frac{
  \mathcal{K}(\Phi_{{\bf m},\sigma^2},\pi) + \log\left(\frac{2}{\varepsilon}\right)}{\lambda} \right\}
  \\
   = \inf_{(m,\sigma^2)}
  \left\{ \int R^H {\rm d}\Phi_{{\bf m},\sigma^2} + \frac{\lambda}{2n}-\frac1{\lambda} \log\left(1-\frac{\vartheta\lambda^2 c_x^2}{4n}\right) + 
  2 \frac{
\sum_{j=1}^{d} \left[
  \frac{1}{2}\log\left(\frac{\vartheta^2}{\sigma^2}\right)
        + \frac{\sigma^2}{\vartheta^2}
  \right]
  + \frac{\|m\|^2}{\vartheta^2} - \frac{d}{2}
 + \log\left(\frac{2}{\varepsilon}\right)}{\lambda} \right\}.
\end{dmath*}
We use the fact that the hinge loss is Lipschitz
and that the $(X_i)$ are uniformly bounded $\Vert X\Vert_\infty<c_x$. 
We get $R^H(\theta)\leq\bar{R}^H+c_x\sqrt{d}\|\theta-\bar{\theta}\|$ and restrict the infemum to distributions
$\nu$ such that $m=\overline{\theta}$:
$$
\mathcal{B}(\mathcal{F}_1)
\leq \overline{R}^H
    + \inf_{\sigma^2}
    \left\{  c_x d \sigma^2
    + \frac{\lambda}{2n}-\frac1{\lambda}\log\left(1-\frac{\vartheta^2\lambda^2 c_x^2}{4n}\right) + \frac{
  d\log\left(\frac{\vartheta^2}{\sigma^2}\right)
        + \frac{2d \sigma^2}{\vartheta^2}
  + \frac{2}{\vartheta^2} - d
 + 2\log\left(\frac{2}{\varepsilon}\right)}{\lambda} \right\}.
$$
We specify $\sigma^2=\frac1{\sqrt{dn}}$ and $\lambda=c_x\sqrt{\frac{n}{\vartheta^2}}$ such that we get:
$$
\mathcal{B}(\mathcal{F}_1)
\leq \overline{R^H}
    + 
    c_x \sqrt{\frac{d}{n}}
    + \frac{\sqrt{\vartheta^2}}{2c_x\sqrt{n}}-c_x\sqrt{\frac{\vartheta^2}{n}}\log\left(1-\frac{1}{4}\right) +
    d\frac{c_x\vartheta}{\sqrt{n}}\log\left(\vartheta^2\sqrt{nd}\right)
        +  c_x\vartheta\frac{\frac{2d}{n\vartheta^2}
  + \frac{2}{\vartheta^2} - d
  + 2\log\left(\frac{2}{\varepsilon}\right)}{\sqrt{n}} .
$$
To get the correct rate we take the prior variance to be $\vartheta^2=\frac1{d}$ by replacing in the above equation we get the desired result.

$\square$

\noindent \textit{Proof of Theorem~\ref{thmconvex}.}
From \cite{Nesterov2004} (th. 3.2.2) we have the following bound on the objective function minimized by VB,
(the objective is not uniformlly Lipschitz)
\begin{equation}
	\label{RateOpt}
	\rho^k(r_n^H)+\frac1\lambda\K(\rho^k,\pi)
	-\inf_{\rho\in\mathcal{F}_1}\left\{\rho(r_n^H)+\frac1\lambda\K(\rho,\pi)\right\}\leq \frac{LM}{\sqrt{1+k}}.
\end{equation}
We have from equation \eqref{pacb1}  specified for measures $\rho^k$ probability $1-\varepsilon$,
\begin{align*}
 \lambda \int r_n^H {\rm d}\rho^k
 &\leq \lambda  \int R^H {\rm d}\rho^k + f(\lambda,n)
  + \mathcal{K}(\rho^k,\pi) + \log\left(\frac{1}{\varepsilon}\right)
\end{align*}
Combining the two equations yields,
\[
  \int R^H {\rm d}\rho^k	\leq  \frac{LM}{\sqrt{1+k}}
  +\frac1\lambda f(n,\lambda)
  +\inf_{\rho\in\mathcal{F}_1}\left\{\rho(r_n^H)
  +\frac1\lambda\K(\rho,\pi)\right\}+\frac1\lambda\log\frac1{\varepsilon}
\]
We can therefore write for any $\rho\in \mathcal{F}_1$,
\[
  \int R^H {\rm d}\rho^k	\leq  \frac{LM}{\sqrt{1+k}}+\frac1\lambda f(n,\lambda)
  +\rho(r_n^H)+\frac1\lambda\K(\rho,\pi)
  +\frac1\lambda\log\frac1{\varepsilon}
\]
Using equation \eqref{pacb1} a second time we get with probability $1-\varepsilon$
\[
  \int R^H {\rm d}\rho^k	\leq  \frac{LM}{\sqrt{1+k}}+\frac2\lambda f(n,\lambda)
  +\rho(R^H)+\frac2\lambda\K(\rho,\pi)
  +\frac2\lambda\log\frac2{\varepsilon}
\]
Because this is true for any $\rho\in\mathcal{F}_1$ in $1-\varepsilon$ we can write the bound for the smallest 
measure in $\mathcal{F}_1$.
\[
  \int R^H {\rm d}\rho^k	\leq  \frac{LM}{\sqrt{1+k}}+\frac2\lambda f(n,\lambda)
  +\inf_{\rho \in \mathcal{F}_1}\left\{\rho(R^H)+\frac2\lambda\K(\rho,\pi)\right\}
  +\frac2\lambda\log\frac2{\varepsilon}
\]
By taking the Gaussian measure with variance $\frac1n$ and mean $\overline{\theta}$ in the infemum and 
taking $\lambda=\frac1{c_x}\sqrt{nd}$ and $\vartheta=\frac1d$, we can use the results of 
Corrolary~\ref{corhinge} to get the result.$\square$

\subsection{Proofs of Section~\ref{sectionranking}}

\noindent \textit{Proof of Lemma~\ref{lemmahoeffding2}.}
The idea of the proof is to use Hoeffding's decomposition of
U-statistics combined with Hoeffding's inequality
for iid random variables.
This was done in ranking by~\cite{Clemencon2008a}, and later
in~\cite{Robbiano2013,Ridgway2014} for ranking via aggregation and
Bayesian statistics. The proof is as follows: we define
$$q^{\theta}_{i,j} = \mathbf{1}_{(Y_i-Y_j)(f_{\theta}(X_i)-f_{\theta}(X_j))<0}
   - R(\theta) $$
so that
$$ U_n :=
 \frac{1}{n(n-1)}\sum_{i,j} q^{\theta}_{i,j} = r_n(\theta)-R(\Theta) .$$
From~\cite{Hoeffding1948} we have
$$
U_n = \frac{1}{n!}\sum_{\pi} \frac{1}{\lfloor \frac{n}{2}\rfloor}
\sum_{i=1}^{\lfloor \frac{n}{2}\rfloor} q^{\theta}_{\pi(i),
\pi(i+\lfloor \frac{n}{2}\rfloor)}
$$
where the sum is taken over all the permutations $\pi$ of $\{1,\dots,n\}$.
Jensen's inequality leads to
\begin{align*}
\mathbb{E} \exp[\lambda U_n]
& = \mathbb{E} \exp\left[\lambda
\frac{1}{n!}\sum_{\pi} \frac{1}{\lfloor \frac{n}{2}\rfloor}
\sum_{i=1}^{\lfloor \frac{n}{2}\rfloor}
q^{\theta}_{\pi(i),\pi(i+\lfloor \frac{n}{2}\rfloor)}
\right] \\
& \leq \frac{1}{n!}\sum_{\pi} \mathbb{E} \exp\left[
\frac{\lambda}{\lfloor \frac{n}{2}\rfloor}
\sum_{i=1}^{\lfloor \frac{n}{2}\rfloor}
q^{\theta}_{\pi(i),\pi(i+\lfloor \frac{n}{2}\rfloor)}
\right].
\end{align*}
We now use, for each of the terms in the sum we use the same argument as in
the proof of Lemma~\ref{lemmahoeffding1} to get
$$
\mathbb{E} \exp[\lambda U_n]
\leq \frac{1}{n!}\sum_{\pi} \exp\left[
\frac{\lambda^2 }{2 \lfloor \frac{n}{2}\rfloor}
\right]
 \leq \exp\left[
\frac{\lambda^2 }{n-1}\right]
$$
(in the last step, we used $\lfloor \frac{n}{2}\rfloor \geq (n-1)/2$).
We proceed in the same way to upper bound $\mathbb{E} \exp[-\lambda U_n]$.
$\square$

\noindent \textit{Proof of Lemma~\ref{lemmabernstein2}.}
As already done above, we use Bernstein inequality and Hoeffding decomposition.
Fix $\theta$. We define this time
$$q^{\theta}_{i,j}=\mathbf{1}\{\left<\theta,X_i - X_j\right>(Y_i-Y_j)<0\} 
- \mathbf{1}\{[\sigma(X_i)-\sigma(X_j)](Y_i-Y_j)<0\} 
- R(\theta)+\overline{R}$$ so that
 $$ U_n := r_n(\theta)-\overline{r}_n - R(\theta)+
\overline{R} = \frac{1}{n(n-1)}\sum_{i\neq j} q^{\theta}_{i,j} .$$
Then,
$$
U_n = \frac{1}{n!}\sum_{\pi} \frac{1}{\lfloor \frac{n}{2}\rfloor}
\sum_{i=1}^{\lfloor \frac{n}{2}\rfloor}
q^{\theta}_{\pi(i),\pi(i+\lfloor \frac{n}{2}\rfloor)}.
$$
Jensen's inequality:
\begin{align*}
\mathbb{E} \exp[\lambda U_n]
& = \mathbb{E} \exp\left[\lambda 
\frac{1}{n!}\sum_{\pi} \frac{1}{\lfloor \frac{n}{2}\rfloor}
\sum_{i=1}^{\lfloor \frac{n}{2}\rfloor} q^{\theta}_{\pi(i),\pi(i+\lfloor \frac{n}{2}\rfloor)}
\right] \\
& \leq \frac{1}{n!}\sum_{\pi} \mathbb{E} \exp\left[
\frac{\lambda}{\lfloor \frac{n}{2}\rfloor}
\sum_{i=1}^{\lfloor \frac{n}{2}\rfloor} q^{\theta}_{\pi(i),\pi(i+\lfloor \frac{n}{2}\rfloor)}
\right].
\end{align*}
Then, for each of the terms in the sum, use Bernstein's inequality:
$$
 \mathbb{E} \exp\left[
\frac{\lambda}{\lfloor \frac{n}{2}\rfloor}
\sum_{i=1}^{\lfloor \frac{n}{2}\rfloor} q^{\theta}_{\pi(i),\pi(i+\lfloor \frac{n}{2}\rfloor)}
\right]
\leq
\exp\left[\frac{
\mathbb{E}((q^{\theta}_{\pi(1),\pi(1+\lfloor \frac{n}{2}\rfloor)})^2)
\frac{\lambda^2}{\lfloor \frac{n}{2}\rfloor}
}{2\left(1-2\frac{\lambda}{\lfloor \frac{n}{2}\rfloor}\right)}\right].
$$
We use again $\lfloor \frac{n}{2}\rfloor\geq (n-1)/2$. Then,
as the pairs $(X_i,Y_i)$ are iid, we have
$\mathbb{E}((q^{\theta}_{\pi(1),\pi(1+\lfloor \frac{n}{2}\rfloor)})^2)
= \mathbb{E}((q^{\theta}_{1,2})^2)$ and then
$\mathbb{E}((q^{\theta}_{1,2})^2) \leq C [R(\theta)-\overline{R}]$ thanks to
the margin assumption. So
$$
 \mathbb{E} \exp\left[
\frac{\lambda}{\lfloor \frac{n}{2}\rfloor}
\sum_{i=1}^{\lfloor \frac{n}{2}\rfloor} q^{\theta}_{\pi(i),\pi(i+\lfloor \frac{n}{2}\rfloor)}
\right]
\leq
\exp\left[\frac{
C [R(\theta)-\overline{R}]
\frac{\lambda^2}{n-1}
}{\left(1-\frac{4\lambda}{n-1}\right)}\right].
$$
This ends the proof of the proposition.
$\square$

\noindent \textit{Proof of Corollary~\ref{corrankinghoeffding}.}
The calculations are similar to the ones in the proof of
Corollary~\ref{corclassifhoeffding} so we don't give the details.
Note that when we reach
$$
\mathcal{B}_{\lambda}(\mathcal{F}_1)
    \leq \overline{R}
    + \frac{2 \lambda}{n-1} + \frac{ c\sqrt{d} 
  + d\log(2\lambda)
 + 2\log\left(\frac{2{\rm e}}{\varepsilon}\right)}{\lambda},
$$
an approximate minimization with respect to $\lambda$ leads to the choice
$\lambda = \sqrt{\frac{d(n-1)}{2}} $.
$\square$

\subsection{Proofs of Section \ref{sectionmatrixcompletion}}

\noindent \textit{Proof.}
First, note that, for any $\rho$,
\begin{align*}
\mathcal{K}(\rho,\pi_{\beta})
& = \beta \int (R-\overline{R}) {\rm d}\rho
         + \mathcal{K}(\rho,\pi) + \log\int
          \exp\left[-\beta(R-\overline{R})\right] {\rm d}\pi
\\
& \leq  \beta \int (R-\overline{R}) {\rm d}\rho
         + \mathcal{K}(\rho,\pi).
\end{align*}
Now, we define a subset of $\mathcal{F}$ that will be used for the
calculation of the bound. We define for $\delta>0$ the probability
distribution
$ \rho_{U,V,\delta}({\rm d}\theta)$
as $\pi$ conditioned to $\theta=\mu \nu^T$ with $\mu$
is uniform on $\{\forall (i,\ell), |\mu_{i,\ell}-U_{i,\ell}|
\leq \delta \}$ and $\nu$
is uniform on $\{\forall (j,\ell), |\nu_{i,\ell}-V_{j,\ell}|
\leq \delta \}$. Note that
\begin{align*}
\int (R-\overline{R}) {\rm d}\rho_{M,N,\delta}
& = \int \mathbb{E}( (\theta_X-M_X)^2 ) \rho_{U,V,\delta}({\rm d}\theta)
\\
& \leq \int 3 \mathbb{E}( ((UV^T)_X-M_X)^2 ) \rho_{U,V,\delta}({\rm d}(\mu,\nu))
\\ & \quad + 3 \int \mathbb{E}( ((U\nu^T)_X-(UV^T)_X)^2 ) \rho_{U,V,\delta}({\rm d}(\mu,\nu))
\\ & \quad + 3 \int \mathbb{E}( ((\mu \nu^T)_X-(U\nu^T)_X)^2 ) \rho_{U,V,\delta}({\rm d}(\mu,\nu)).
\end{align*}
By definition, the first term is $=0$.
Moreover:
\begin{align*}
  \lefteqn{\int \mathbb{E}( ((U\nu^T)_X-(UV^T)_X)^2 ) \rho_{U,V,\delta}({\rm d}(\mu,\nu))} \quad  \\
  & = \int \frac{1}{m_1 m_2}\sum_{i,j}
        \left[ \sum_k U_{i,k}(\nu_{j,k}-V_{j,k}) \right]^2
  \rho_{U,V,\delta}({\rm d}(\mu,\nu))
  \\
  & \leq   \int \frac{1}{m_1 m_2}\sum_{i,j}
        \left[\sum_k U_{i,k}^2 \right] \left[\sum_k (\nu_{j,k}-V_{j,k})^2\right]
  \rho_{U,V,\delta}({\rm d}(\mu,\nu))
  \\
  & \leq K r C^2 \delta^2.
\end{align*}
In the same way,
\begin{align*}
  \int \mathbb{E}( ((\mu \nu^T)_X-(U \nu^T)_X)^2 ) \rho_{U,V,\delta}({\rm d}(\mu,\nu))
 & \leq \int \|\mu-U\|_F^2 \|\nu\|_F^2
  \rho_{U,V,\delta}({\rm d}(\mu,\nu))
\\
& \leq K r (C+\delta)^2 \delta^2.
\end{align*}
So:
$$
\int (R-\overline{R}) {\rm d}\rho_{M,N,\delta}
 \leq 2 K r \delta^2 (C+\delta^2).
$$
Now, let us consider the term $\mathcal{K}(\rho_{U,V,\delta},\pi)$.
An explicit calculation is possible but tedious. Instead, we might just
introduce the set $\mathcal{G}_{\delta}=\{\theta = \mu \nu^T, \|\mu-U\|_F \leq \delta,
\|\nu-V\|_F \leq \delta \}$ and note that $\mathcal{K}(\rho_{U,V,\delta},\pi)
\leq \log \frac{1}{\pi(\mathcal{G}_{\delta})}$.
An upper bound for $\mathcal{G}_{\delta}$ is calculated
page 317-320 in~\cite{Alquier2014} and the result is given by~(10) in this
reference:
\begin{multline*}
 \mathcal{K}(\rho_{U,V,\delta},\pi)
 \leq 4\delta^2 + 2\|U\|_F^2 + 2 \|N\|_F^2 + 2\log(2)
   \\
   + (m_1+m_2) r \log\left(\frac{1}{\delta}\sqrt{\frac{3\pi (m_1\vee m_2)K}{4}} \right)
    + 2K \log\left(\frac{\Gamma(a)3^{a+1} \exp(2)}{b^{a+1}2^a }\right)
\end{multline*}
as soon as the restriction $b\leq \frac{\delta^2}{2 m_1 K \log(2 m_1 K)},
\frac{\delta^2}{2 m_2 K \log(2 m_2 K)}$ is satisfied. So we obtain:
\begin{multline*}
 \mathcal{K}(\rho_{U,V,\delta},\pi_{\beta})
 \leq \beta 2 K r \delta^2 (C+\delta^2)
 + 4\delta^2 + 2\|U\|_F^2 + 2 \|N\|_F^2 + 2\log(2)
   \\
   + (m_1+m_2) r \log\left(\frac{1}{\delta}\sqrt{\frac{3\pi (m_1\vee m_2)K}{4}} \right)
    + 2K \log\left(\frac{\Gamma(a)3^{a+1} \exp(2)}{b^{a+1}2^a }\right).
\end{multline*}
Note that $\|U\|_F^2 \leq C^2 r m_1$, $\|V\|_F^2 \leq C^2 r m_2 $ and
$K\leq m_1 + m_2$ so it is clear
that the choice $\delta = \sqrt{\frac{1}{\beta}}$ and $b\leq
\frac{1}{2\beta (m_1\vee m_2) \log(2 K (m_1\vee m_2))}$ leads to the existence of
a constant $\mathcal{C}(a,C)$ such that
$$
\mathcal{K}(\rho_{U,V,\delta},\pi_{\beta})
\leq
\mathcal{C}(a,C) \left\{ r(m_1 + m_2 ) \log\left[\beta b (m_1+m_2)K\right]
 +\frac{1}{\beta} \right\}.
$$

$\square$

\section{Implementation details}

\subsection{Sequential Monte Carlo}
\label{app:SMC}

Tempering SMC approximates iteratively a sequence of distribution $\rho_{\lambda_t}$, with
\[
	\rho_{\lambda_t}(\dd\theta) = \frac{1}{Z_t}
	\exp\left(-\lambda_t r_n(\theta)\right)\pi(\dd\theta),
\]
and temperature ladder $\lambda_0=0<\ldots<\lambda_T=\lambda$. The pseudo code below is given for an adaptive sequence of temperatures.

\begin{algorithm}[H]
\caption{Tempering SMC}\label{algo-smc}
\begin{description}
\item[Input] $N$ (number of particles),  $\tau\in(0,1)$ (ESS threshold), $\kappa>0$ (random walk tuning parameter)
\item[Init.] Sample $\theta_0^i\sim\pi_\xi(\theta)$ for $i=1$ to $N$, set $t\leftarrow 1$, $\lambda_0=0$, $Z_0=1$. 
\item[Loop]  
\begin{description}
\item[a.] Solve in $\lambda_t$ the equation
\begin{equation}\label{eq:ESS}
\frac{\{\sum_{i=1}^N w_t(\theta_{t-1}^i)\}^2}
{\sum_{i=1}^N \{w_t(\theta_{t-1}^i))^2\} } = \tau N, 
\quad w_t(\theta) = \exp[ -(\lambda_t-\lambda_{t-1}) r_n(\theta) ]
\end{equation}
using bisection search. If $\lambda_t\geq\lambda_T$, set 
$Z_T=Z_{t-1}\times \left\{\frac{1}{N} \sum_{i=1}^N w_t(\theta_{t-1}^i)\right\}$, and stop. 
\item[b.] Resample: for $i=1$ to $N$, draw $A_t^i$ in $1,\ldots,N$ so that 
	$\P(A_t^i=j) = w_t(\theta_{t-1}^j)/\sum_{k=1}^N w_t(\theta_{t-1}^k)$; see Algorithm \ref{algo-sys} 
in the appendix. 
\item[c.] Sample $\theta_t^i \sim M_t(\theta_{t-1}^{A_t^i},\dd \theta)$ for $i=1$ to $N$
 where $M_t$ is a MCMC kernel that leaves invariant $\pi_t$; see comments below. 
\item[d.] Set $Z_t = Z_{t-1}\times \left\{ \frac{1}{N} \sum_{i=1}^N w_t(\theta_{t-1}^i) \right\}$.
\end{description}
\end{description}
\end{algorithm}

The algorithm outputs a weighted sample $(w^i_T,\theta_T^i)$ approximately distributed as 
target posterior, and an unbiased estimator of the normalizing constant $Z_{\lambda_T}$.

Step {\bf b.} of algorithm \ref{app:SMC} depends of a resampling algorithm. We choose to use 
Systematic resampling, described in Algorithm \ref{algo-sys}.
\begin{algorithm}[H]
\caption{Systematic resampling
\label{algo-sys}}
\begin{description}
\item[Input:] Normalised weights 
$W_t^j\eqdef w_t(\theta_{t-1}^j)/\sum_{i=1}^N w_t(\theta_{t-1}^i)$.
\item[Output:] indices $A^i\in\{1,\ldots,N\}$, for $i=1,\ldots,N$. 
\item[a.] Sample $U\sim \Unif$.
\item[b.] Compute cumulative weights as 
$C^n=\sum_{m=1}^n NW^m$.
\item[c.] Set $s\leftarrow U$, $m\leftarrow 1$. 
\item[d.] \textbf{For} $n= 1:N$
\item[]$\qquad$ \textbf{While} $C^m<s$ \textbf{do} $m\leftarrow m+1$. 
\item[]$\qquad$ $A^n\leftarrow m$, and $s\leftarrow s+1$. 
\item[]$\quad$ \textbf{End For}
\end{description}
\end{algorithm}

For the MCMC step, we used a Gaussian random-walk Metropolis kernel, with a covariance
matrix for the random step that is proportional to the empirical covariance matrix
of the current set of simulations.

\subsection{Optimizing the bound}
\label{sec:DA}

A natural idea to find a global optimum of the objective is to try to solve a sequence 
of local optimization problems with increasing temperatures. For $\gamma=0$ the problem can be solved exactly
(as a KL divergence between two Gaussians). Then, for two consecutive temperatures,  the corresponding solutions should be close enough. 




This idea has been coined under several names. It has a long history in variational algorithm under the name
deterministic annealing, \cite{Yuille} uses it on mean field on Gibbs distribution 
for Markov random fields. 
In addition the intermediate results can be of interest in our case for selecting the temperature. One can 
compute the bound at almost no additional cost as a function of the current risk. In turns this can be used to 
monitor the bound.

\begin{algorithm}[H]
\caption{Deterministic annealing}\label{algo-cont}
\begin{description}
	\item[Input] $(\lambda_t)_{t\in[0,T]}$ a sequence of temperature
\item[Init.] Set $m=0$ and $\Sigma=\vartheta I_d$, the values minimizing KL-divergence for $\lambda=0$  
\item[Loop] t=1,\dots,T
\begin{description}
	\item[a.] $m^{\lambda_t},\Sigma^{\lambda_t}$ = Minimize $\mathcal{L}^{\lambda_t}(m,\Sigma)$ using some local optimization routine with initial points $m^{\lambda_{t-1}},\Sigma^{\lambda_{t-1}}$
  \item[b.] Break if the empirical bound increases.
\end{description}
\item[End Loop]
\end{description}
\end{algorithm}

\begin{figure}[H]
\begin{center}
\begin{tabular}{ll}
\subfloat[A one dimensional problem]{\includegraphics[scale=0.4]{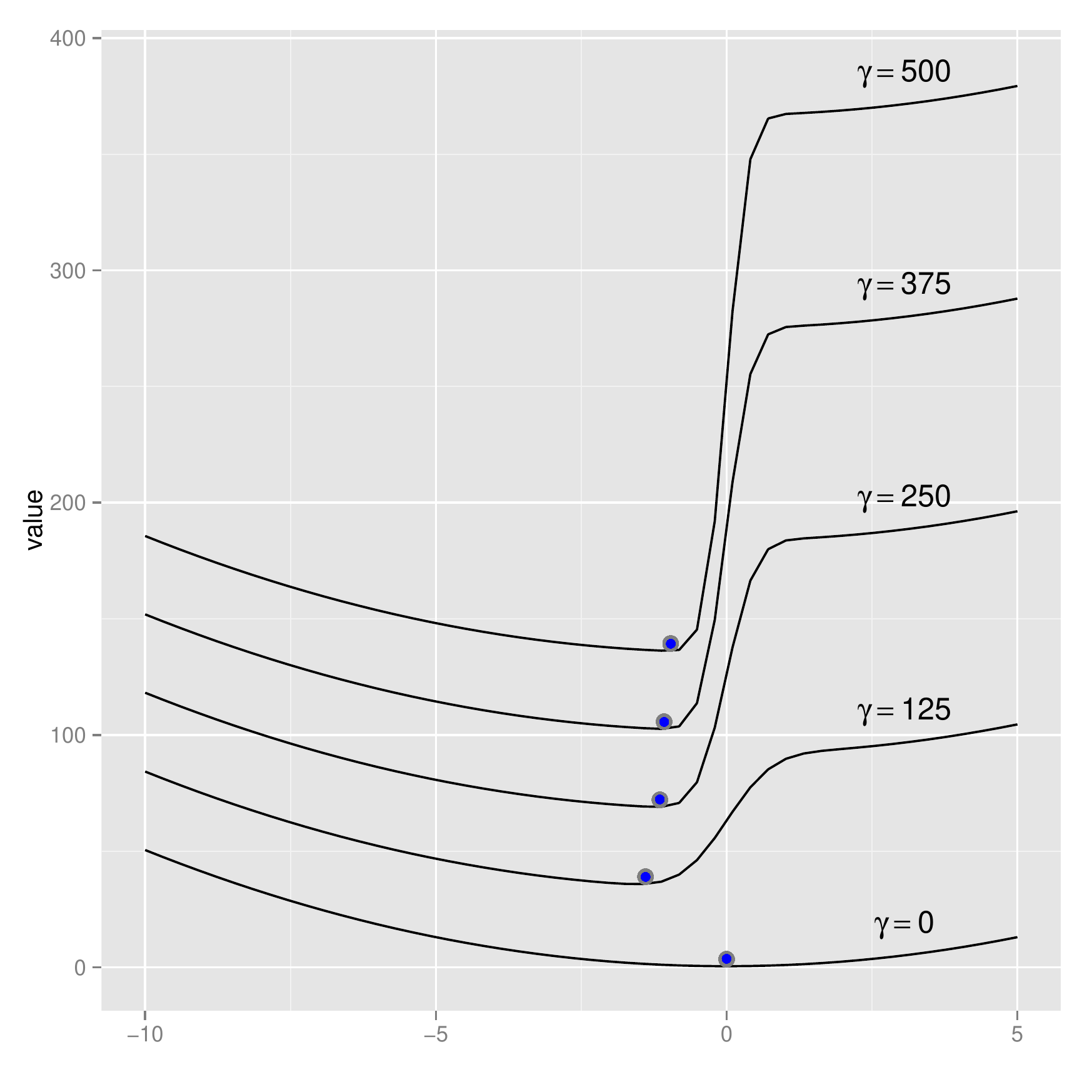}} &
\subfloat[Empirical bound]{\includegraphics[scale=0.4]{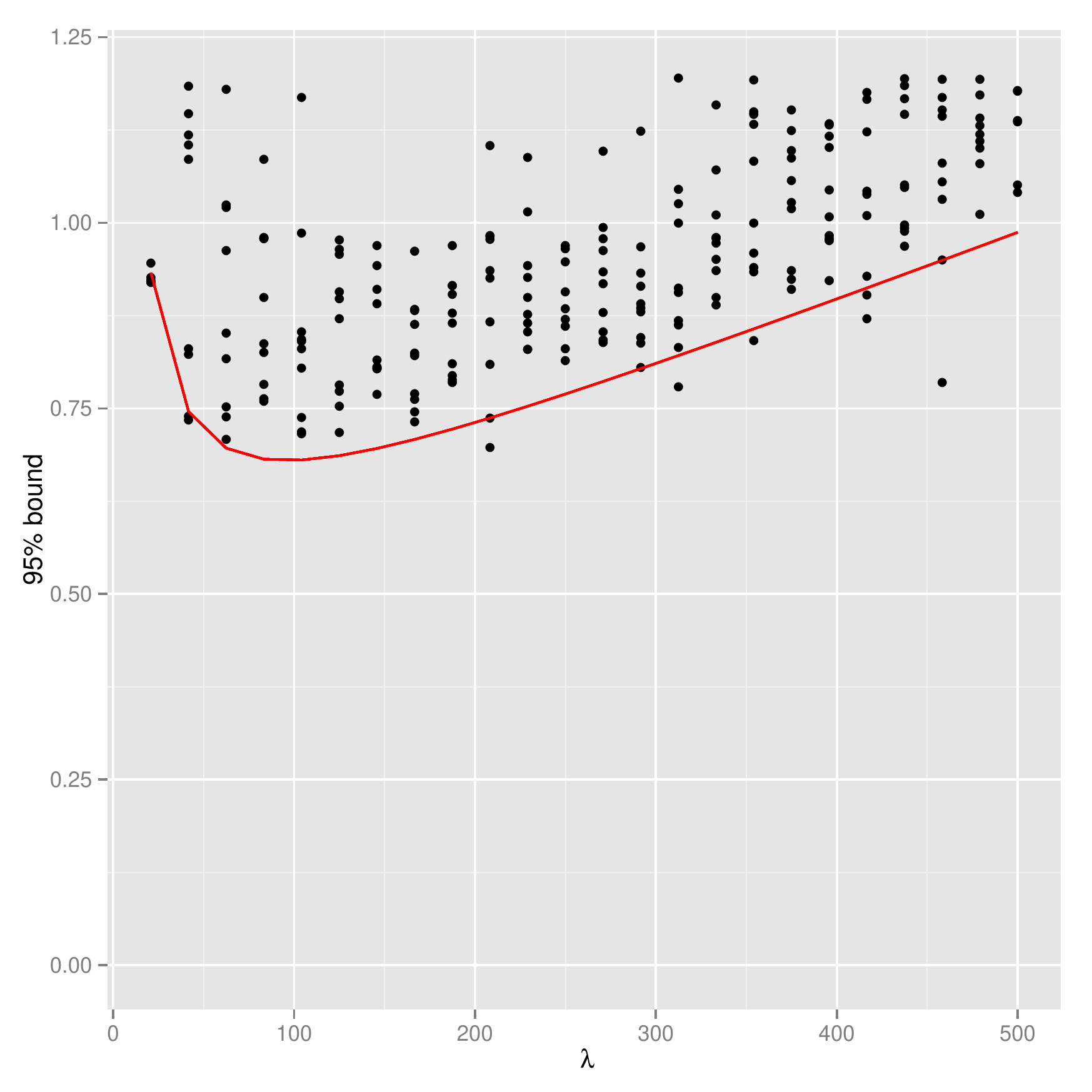} } 
\end{tabular}
\caption{Deterministic annealing on a Pima Indians with one covariate and full model resp.}
\label{fig:Cont}
\begin{minipage}{15cm}
\footnotesize{The right panel gives the empirical bound obtained for the DA method (in red)  and 
the dot are direct global optimization based on L-BFGS algorithms from starting values drawn from
the prior. Each optimization problem is repeated $20$ times.}
\vspace*{3mm}
\hrule
\end{minipage}
\end{center}
\end{figure}

We find that using a deterministic annealing algorithm with a limited amount of steps helps in finding a high enough optimum. 
On the left panel of Figure \ref{fig:Cont}, we can see the one dimensional case where the initial problem 
$\gamma=0$ corresponds to a convex minimization problem and where the increasing temperature 
gradually complexifies the optimization problem. Figure \ref{fig:Cont} shows that the solution 
given by DA is in average lower than randomly initialized optimization.

\section{Stochastic gradient descent}

The stochastic gradient descent algorithm used in Section \ref{sec:ranking} is described as Algorithm \ref{algo-SGD}.

\begin{algorithm}[H]
\caption{Stochastic Gradient Descent}\label{algo-SGD}
\begin{description}
	\item[Input] $B$ a batch size, an unbiased estimator of the gradient $\hat{\nabla}_B f$, $\eta\in (0,1)$ and $c$ 
	\item[While]$\lnot\text{converged}$ 
\begin{description}
	\item[a.] $x_{t+1}=x_t-\lambda_t \hat{\nabla}_B f(x_t)$
	\item[b.] Update $\lambda_{t+1}=\frac1{(t+c)^\eta}$	
\end{description}
\item[End Loop]
\end{description}
\hrule
\vspace*{3mm}
\begin{minipage}{15cm}
In all our experiment we take $c=1$ and $\eta=0.9$. 
\vspace*{3mm}
\hrule
\end{minipage}
\end{algorithm}
\end{document}